\documentclass[runningheads]{llncs}
\usepackage[pdftex]{graphicx}
\usepackage{epstopdf}
\usepackage{amsfonts}
 \usepackage{graphicx}
 \usepackage[scriptsize]{subfigure}
 \usepackage{color}
 \usepackage{amsmath}
\usepackage{wrapfig}
\usepackage[ruled,linesnumbered,procnumbered]{algorithm2e}
\SetKwInOut{Parameter}{Parameter}
\SetKwRepeat{Do}{do}{while}
\usepackage[hyphens]{url}
\usepackage{microtype}
\usepackage[misc]{ifsym}

\begin{document}
\title{Confusable Learning for Large-class \\Few-Shot Classification\thanks{Preliminary work was done during an internship at Hong Kong Baptist University (HKBU). This research was partially funded by the Australian Government through the Australian Research Council (ARC) under grants LP180100654, HKBU Tier-1 Start-up Grant and HKBU CSD Start-up Grant.}}
%
%
\author{Bingcong Li\inst{1,2} \and
Bo Han\inst{2} \and
Zhuowei Wang\inst{3} \and
Jing Jiang\inst{3} \and
Guodong Long\inst{3}}
\authorrunning{B. Li et al.}
%
\institute{School of Automation, Guangdong University of Technology, China \and
Department of Computer Science, Hong Kong Baptist University, HKSAR, China \and
Australian Artificial Intelligence Institute, University of Technology Sydney, Australia
}

\toctitle{Confusable Learning for Large-class Few-Shot Classification}
\tocauthor{Bingcong~Li, Bo~Han, Zhuowei~Wang, Jing~Jiang, Guodong~Long}
\maketitle              
\setcounter{footnote}{0}
\begin{abstract}
Few-shot image classification is challenging due to the lack of ample samples in each class. Such a challenge becomes even tougher when the number of classes is very large, i.e., the large-class few-shot scenario. In this novel scenario, existing approaches do not perform well because they ignore confusable classes, namely similar classes that are difficult to distinguish from each other. These classes carry more information. In this paper, we propose a biased learning paradigm called \textit{Confusable Learning}, which focuses more on confusable classes. Our method can be applied to mainstream meta-learning algorithms. Specifically, our method maintains a dynamically updating confusion matrix, which analyzes confusable classes in the dataset. Such a confusion matrix helps meta learners to emphasize on confusable classes. Comprehensive experiments on \textit{Omniglot}, \textit{Fungi}, and \textit{ImageNet} demonstrate the efficacy of our method over state-of-the-art baselines.
\keywords{Large-Class Few-Shot Classification  \and Meta-Learning \and Confusion Matrix.}
\end{abstract}

\section{Introduction}

Deep Learning has made significant progress in many areas recently, but it relies on numerous labeled instances. Without enough labeled instances, deep models usually suffer from severe over-fitting, while a human can easily learn patterns from a few instances. By incorporating this ability, meta-learning based few-shot learning has become a hot topic~\cite{finn2017model,snell2017prototypical}. Mainstream meta-learning methods obtain meta-knowledge from a base dataset containing a large number of labeled instances, and employ meta-knowledge to classify an meta-testing dataset.

Recent progress in few-shot learning focuses on small-class few-shot scenario~\cite{wang2019few}. These methods consist of three directions: model initialization based methods, metric learning methods and data augmentation. In particular, model initialization based methods include learning a good model initialization~\cite{finn2017model}. Metric learning methods assume that there exists an embedding for any given dataset, where the representation of instances drawn from the same class is close to each other~\cite{vinyals2016matching,snell2017prototypical}. The predictions of these methods are conditioned on distance or metric to few labeled instances during meta-training. Data augmentation generates new instances based on existing ``seed'' instances~\cite{chen2018a}.

\begin{table}
	\centering
	\caption{Difficulties of different learning scenarios.}
	\label{table:com}
	\scalebox{1}
	{
	\scalebox{1}{
	
		\begin{tabular}{c | c | c}
			\hline
			& few-shot & many-shot \\ \hline
			small-class & hard ~\cite{finn2017model} & easy ~\cite{he2016deep}\\ \hline
			large-class & \textbf{hardest}~\cite{Li_2019_CVPR} & medium ~\cite{deng2010does}\\ \hline
		\end{tabular}
		}
	}
\end{table}

Most existing methods are evaluated on tasks with less than $50$ classes \cite{NIPS2018_7504,NIPS2018_7352,liu2018learning,pmlr-v80-franceschi18a}. However, in practice, we are often asked to classify thousands of classes, which naturally brings large-class few-shot scenario~\cite{hariharan2017low,Li_2019_CVPR}. As shown in Table~\ref{table:com}, this scenario is challenging to conquer. In this kind of scenario, some classes are more difficult for the model to classify. Performance of the model on these classes will suffer from a relatively low accuracy. Meanwhile, experiments in large-class many-shot scenario also provides the same conclusion~\cite{deng2010does}.

\begin{figure}[t]
\centering
\label{fig:fig_motivation:1}
\includegraphics[width=8cm]{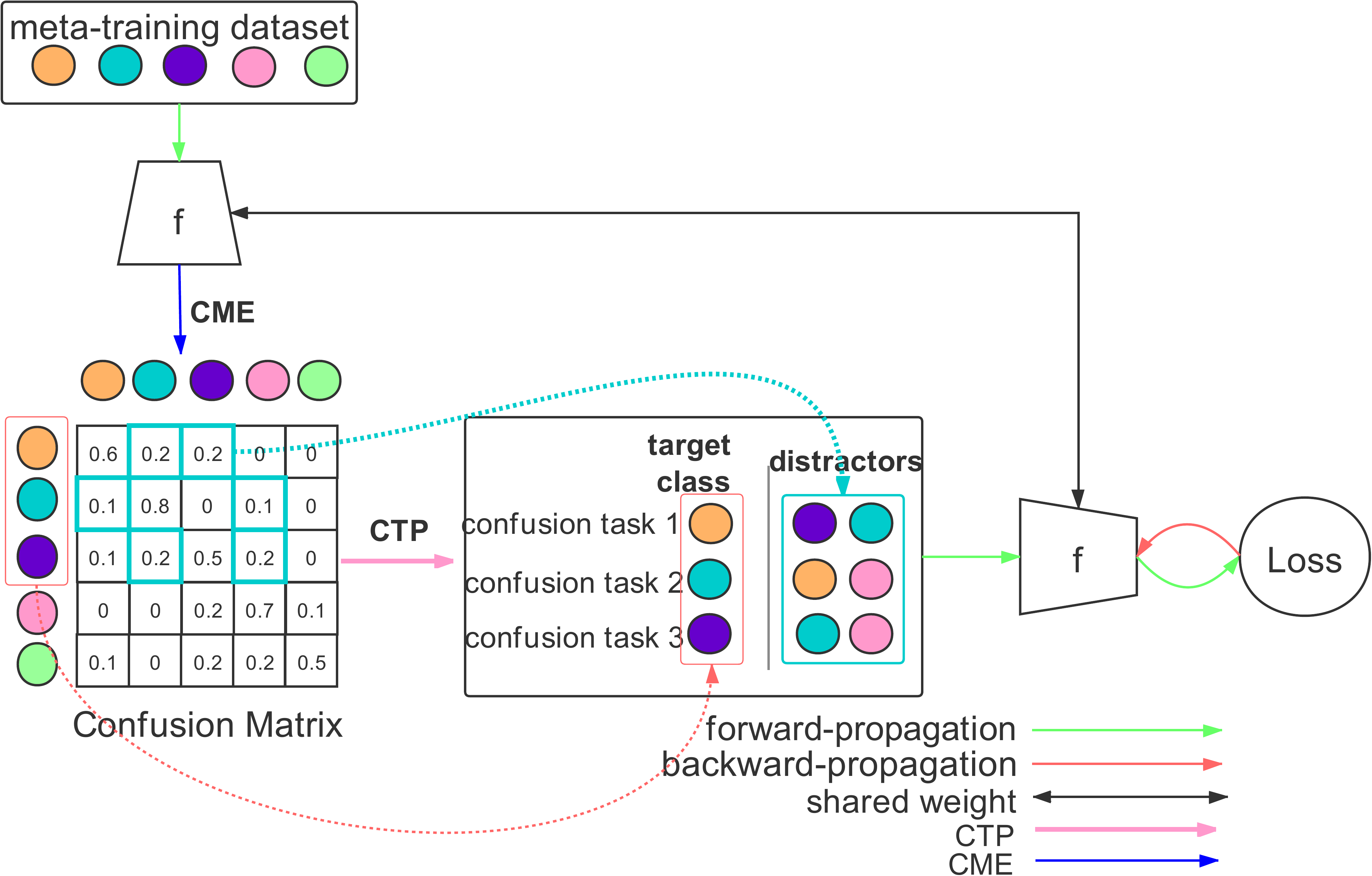} 
\caption{Demonstration of one episode of \textit{Confusable Learning} as specified in Algorithm~\ref{alg:algorithm1}. In this example, there are five classes in total in the meta-training dataset, which are represented by five circles with different colors. In the Confusion Matrix, the entry $\textbf{C}_{i,j}$ (i$\neq$j) is the average probability for instances of class $i$ to be misclassified as class $j$. $f$ denotes the meta-learning algorithm, where \textit{Confusable Learning} is applied. Specifically, we use \textit{Confusion Task Processor} in Algorithm~\ref{alg:algorithm3} to sample three classes (red box) as our \textit{target classes} and then locate two \textit{distractors} for each target class (grey box). For example, for the orange class, we choose the purple and blue classes in the grey box as its \textit{distractors}, because they have a higher probability in the Confusion Matrix. Second, \textit{Confusion Matrix Estimation (CME)} in Algorithm~\ref{alg:algorithm2}, represented by the blue arrow, is applied to update the Confusion Matrix using probabilities calculated by the meta-learner. Confusable classes change as the model updates.} 
\label{fig:fig demo}
\end{figure}

To tackle the large-class few-shot scenario, we propose a biased learning paradigm called \textit{Confusable Learning}. Our key idea is to focus on confusable classes in meta-training dataset, which can improve model robustness in meta-testing dataset. 
In each iteration, we uniformly sample a few classes and denote them as \textit{target classes}. For each target class, our paradigm selects several similar classes, which the model has difficulty in distinguishing from their target class. We call these classes \textit{distractors}\footnote{Distractors defined in our paper are different from those defined by Ren~\cite{Ren2018Meta}.}. The model is then trained by a meta-learning algorithm to recognize instances of target class from those of distractors. Note that distractors are dynamically changing: when the model fits the distractors in each iteration, they become less confusable; while other classes become relatively more confusable and have higher chance to be selected as distractors. In this way, the model goes through every class in meta-training dataset dynamically. We briefly show how \textit{Confusable Learning} works in Figure~\ref{fig:fig demo}, where \textit{Confusable Learning} is presented as a framework agnostic to different meta-learners. In the experiment, we build our method on the top of several state-of-the-art meta-learning methods, including Prototypical Network, Matching Network, Prototypical Matching Network and Ridge Regression Differentiable Discriminator~\cite{vinyals2016matching,snell2017prototypical,wang2018low,bertinetto2018metalearning}. \textit{Confusable Learning} is a training framework applied only in meta-training stage. In meta-test stage, these models are evaluated in the same way their authors originally did~\cite{vinyals2016matching,snell2017prototypical,wang2018low,bertinetto2018metalearning}. We evaluate our method on datasets that have more than a thousand classes, including \textit{Omniglot}~\cite{lake2015human}, \textit{Fungi}\footnote{\url{https://github.com/visipedia/fgvcx_fungi_comp}}, and \textit{ImageNet}~\cite{imagenet_cvpr09}. The empirical results show that the models with \textit{Confusable Learning} has better generalization in unseen meta-testing datasets than the models without \textit{Confusable Learning} on large-class few-shot tasks.

\section{Related Literature} 
In the following, we discuss representative few-shot learning algorithms organized into three main categories: model initialization and learnable optimizer based methods, metric learning methods, and data augmentation. The first category is learning an optimizer or a specified initialization. For example, Wichrowska~\textit{et al.}~\cite{wichrowska2017learned} used an RNN network to replace the gradient descent optimizer. Finn~\textit{et al.}~\cite{finn2017model} proposed a gradient-based method to backward-propagate through the learning process of a task, which finds a good initialization for the meta-learner. This method takes a very illuminating perspective on learning which is very effective. However, as reported by Chen~\textit{et al.}~\cite{chen2018a}, this method can be easily hindered by large shifts between training and testing domain.

Metric learning methods assume that there exists an embedding for any given dataset, where the representation of instances drawn from the same class is close to each other. 
For example, Koch~\textit{et al.}~\cite{koch2015siamese} addressed one-shot problem by comparing distances from embedding of query instances to a single support instance. Vinyals~\textit{et al.}~\cite{vinyals2016matching} proposed to focus attention on features that help solve a particular task. Instead of comparing distances to the embedding of labeled instances, Snell~\textit{et al.}~\cite{snell2017prototypical} compared the distance to the prototype of each class, which is computed by averaging the embedding of support instances belonging to that class. In contrast with previous methods using handcraft metrics like Euclidean distance and cosine similarity, Garcia and Bruna~\cite{garcia2018fewshot} used a Graph Neural Network to learn the metric specific to a given~ dataset. Liu~\textit{et al.}~\cite{liu2019prototype,liu2019learning} proposed a novel graph structure to tackle the similar problem.

Data augmentation methods address few-shot problem by generating new instances based on existing ``seed'' instances. Chen~\textit{et al.}~\cite{chen2018a} showed that applying traditional data augmentation methods like crop, flip, and color jitter in traditional many-shot scenario can produce a competitive result. To capture more realistic variation for generating new instances, Hariharan and Girshick~\cite{hariharan2017low} learned to transfer the variation of the meta-training dataset to the meta-testing dataset. Instead of concerning the authenticity of those generated instances, Wang~\textit{et al.}~\cite{wang2018low} proposed to use a hallucinator to produce additional training instances. Data augmentation methods are orthogonal with other few-shot methods and can be considered as the pre-processing procedure for other few-shot methods. Thus, we do not consider these methods in this work.

On the other hand, researchers have studied large-class problem under many-shot and few-shot setting. Deng~\textit{et al.}~\cite{deng2010does} studied the ability of deep models to classify ImageNet subsets with more than 10000 classes of images. They found that only a small portions of classes are truly confusable and suggested to focus on these confusable classes to improve performance. Gupta~\textit{et al.}~\cite{Gupta2014Training} treated confusable classes as the noise of lower priority and prevented the model from updating classes that are consistently misclassified. However, the assumption of ignoring confusable classes will cause performance degradation. Li~\textit{et al.}~\cite{Li_2019_CVPR} proposed a novel large-class few-shot learning model by learning transferable visual features with the class hierarchy, which encodes the semantic relations between source and target classes. These works admit the difficulty brought by large-class setting. Deng~\textit{et al.}~\cite{deng2010does} motivates us to focus on, instead of ignoring, confusable classes in large-class setting to improve classification performance. Meanwhile, it is necessary to point out that Deng~\textit{et al.}~\cite{deng2010does} did not propose a method to deal with confusable classes, neither did they study how the learning of confusable classes of meta-training dataset influences the model performance on meta-testing~ dataset. 

\SetKwFunction{ConfusionTaskprocessor}{ConfusionTaskprocessor}
\SetKwFunction{ConfusionMatrixEstimation}{ConfusionMatrixEstimation}
\begin{algorithm}[tb]
\caption{\textbf{Confusable Learning}}
\label{alg:algorithm1}
\KwIn{Training set $\mathbb{D}=\{(\textbf{x}^1, y^1), ..., (\textbf{x}^N, y^N)\}$ (denote $\mathbb{D}_k$ as a set containing all instances of $\mathbb{D}$ that belong to $k$th class); Learnable weights $\Theta$; Number of times performing CME $M$.}
\Parameter{$\rho$}
  Initialize confusion matrix $\textbf{E}$ as a matrix of shape $(K, K)$, with each entry initialized as $1/K$\;
 
 \While{Model does not converge}{
    $\textbf{C} \leftarrow \textbf{E}$\;
    \tcp{Algorithm~\ref{alg:algorithm3}}
    $\Theta \leftarrow$ \ConfusionTaskprocessor{$\mathbb{D}$, $\Theta$, $\textbf{C}$}\;
    
  \ForEach{$i$ in $\{1,...,M\}$}{
  \tcp{Algorithm~\ref{alg:algorithm2}}
    $\textbf{E} \leftarrow $ \ConfusionMatrixEstimation{$\mathbb{D}$, $\Theta$, $\textbf{E}$}\;
  }
  }
\end{algorithm}

\section{Algorithm}


Here we clarify the notations that will be used later. Given a meta-training dataset with $K$ classes called $\mathbb{D}=\{(\textbf{x}^1, y^1), ..., (\textbf{x}^N, y^N)\}$, we denote $\mathbb{D}_k$ as a set containing all instances within $\mathbb{D}$ that belong to the $k$th class. For conventional meta-learning methods, a few classes, represented by their indices in our work, are drawn from $K$ classes in each episode. A support set and a held-out query set are sampled from instances of these classes in $\mathbb{D}$.

To focus on confusable classes, we propose our method called \textit{Confusable Learning}. 
Specifically, \textit{Confusable Learning} utilizes the confusion matrix $\textbf{C}$, which is designed as a square matrix of $K$ rows and $K$ columns for a meta-training dataset with $K$ classes. This matrix is calculated by training a model and then setting the entry $\textbf{C}_{i,j}$ as the count of the test instances of class $i$ that are misclassified as class $j$  ~\cite{inproceedings}, as formally shown below:
\begin{equation}
\label{equ:cm}
    \textbf{C}_{i,j} = \sum_{(\textbf{x}, y) \in \mathbb{Q}^i}\textbf{1}_{j=\mathop{\arg\max}_{k}P(\hat{y} = k | \textbf{x})},
\end{equation}
in which $\mathbb{Q}$ denotes a query set, and $P(\hat{y} = k | \textbf{x})$ denotes the prediction given by a meta-learning model.

The confusion matrix is often used to describe the performance of a classification model on a test dataset with true labels. 
We make use of the confusion matrix to find out the confusable classes and then learn from these confusable classes to update parameters in the model. We need to calculate this matrix in each episode because when parameters in the model change, the corresponding confusion matrix changes as well.

In the case where classes are imbalanced, it is useful to normalize the confusion matrix by dividing each entry of the confusion matrix by the sum of the entries of the corresponding row~\cite{normalizedconfusionmatrix}. The normalized confusion matrix $\textbf{C}^{\text{n}}$ is formally given by:
\begin{equation}
\label{equ:normalized_cm}
\textbf{C}^{\text{n}}_{i, j} = \frac{\textbf{C}_{i, j}}{\sum_{k=1}^{K}\textbf{C}_{i, k}}.
\end{equation}
As a result, the sum of entries in each row of $\textbf{C}^{\text{n}}$ is equal to 1.


To learn more about the confusable classes, we propose the definition of soft confusion matrix, which is slightly different from the traditional one. Instead of counting the times of classifying class $i$ as class $j$, we utilize the probability of classifying instances of class $i$ as class $j$:
\begin{equation}
    \label{equ:prob_confusion_matrix}
    \textbf{C}^{\text{p}}_{i,j} = \frac{1}{|\mathbb{Q}^i|} \sum_{(\textbf{x}, y) \in \mathbb{Q}^i} P(\hat{y} = j | \textbf{x}).
\end{equation}
 It is easy to observe that $\sum_{j=1}^{K}\textbf{C}^{\text{p}}_{i,j}$ is always equal to 1. Thus, the soft confusion matrix defined by Eq.~(\ref{equ:prob_confusion_matrix})~is normalized itself. We will show in the experiment section that by using such a definition \textit{Confusable Learning} exploit more detailed information about the error the model made for each instance.

We show the framework of our method in Algorithm~\ref{alg:algorithm1}. Algorithm~\ref{alg:algorithm1} shows that \textit{Confusable Learning} consists of 2 steps: Algorithm \ref{alg:algorithm3} and Algorithm \ref{alg:algorithm2}, which will be introduced in detail in Section \ref{sec:3.1} and Section \ref{sec:3.2} respectively.

\SetKwFunction{RandomSample}{RandomSample}
\begin{algorithm}[tb]
\caption{ConfusionTaskProcessor($\mathbb{D}$, $\Theta$,$\textbf{C}$)}
\KwIn{Confusion Matrix $\textbf{C}$; Training set $\mathbb{D}$; Learnable weights $\Theta$.}
\Parameter{Number of support instances for each class $N_S$; Number of query instances for each class $N_Q$; Number of distractors for each target class $N_D$; Number of confusable tasks $N_{T}^{\text{c}}$.}
\label{alg:algorithm3}
  
  $\bar{v}^{\text{c}} \leftarrow $\RandomSample{$\{1,...,K\}, N_{T}^{\text{c}}$} \tcp*{Sample target classes}
  
  \ForEach{$k$ in $\{\bar{v}^{\text{c}}_1, ..., \bar{v}^{\text{c}}_{N_{T}^{\text{c}}}\}$}{
   Sample distractors $\bar{w}^k$ with Eq.~(\ref{equ:multinomial})\;
   
   \ForEach{$l$ in $\{\bar{w}^k_1, ..., \bar{w}^k_{N_D}\}$}{
    
    \tcp{Sample instances for distractors}
    $\mathbb{S}^{\text{distractor}}_{k,l} \leftarrow $\RandomSample{$\mathbb{D}_{l}, N_S$}\;
   }
   $\mathbb{S}^{\text{distractor}}_{k} \leftarrow \bigcup_{l=\{\bar{w}^k_1, ..., \bar{w}^k_{N_D}\}}\mathbb{S}^{\text{distractor}}_{k,l}$\;
   
   
   $\mathbb{S}^{\text{target}}_{k} \leftarrow $\RandomSample{$\mathbb{D}_{k}, N_S$}\tcp*{Sample instances for target class}

   $\mathbb{Q}_{k} \leftarrow $\RandomSample{$\mathbb{D}_{k}, N_Q$} \tcp*{Sample instances for query set}
  }
 
  $J \leftarrow 0$\;
  
  \ForEach{$k$ in $\{\bar{v}^{\text{c}}_1, ..., \bar{v}^{\text{c}}_{N_{T}^{\text{c}}}\}$}{
   \ForEach{$(\textbf{x},y)$ in $\mathbb{Q}_{k}$}{
    
     Calulate loss for any base meta-learning algorithm using the support set given by $\mathbb{S}^{\text{distractor}}_{k}\cup\mathbb{S}^{\text{target}}_{k}$\;
    $J \leftarrow J + \frac{1}{N_{T}^{\text{c}} N_Q}\log{\text{P}(\hat{y}=k|\textbf{x})}$\;

   }
  }
  Update $\Theta$ by back-propagating on $J$\;
\end{algorithm}

\subsection{Confusion Task Processor}
\label{sec:3.1}
Before delving into the calculation of confusion matrix, we assume that we have a confusion matrix at the start of each episode. Next, \textit{Confusion Learning} constructs several tasks called confusion tasks according to the confusion matrix. Then the meta-learning method is trained on these tasks.

As shown in Algorithm~\ref{alg:algorithm3}, to construct the confusion tasks, we first uniformly sample $N_{T}^{\text{c}}$ target classes $\bar{v}^{\text{c}} = \{\bar{v}^{\text{c}}_1, ..., \bar{v}^{\text{c}}_{N_{T}^{\text{c}}}\}$. For target class $k \in \bar{v}^{\text{c}}$, we use a multinomial distribution to sample $N_D$ classes as distractors $\bar{w}^k = \{\bar{w}^k_1, ..., \bar{w}^k_{N_D}\}$. That is:
\begin{equation}
\label{equ:multinomial}
\bar{w}^k \sim \text{Multinomial}(N_D, \textbf{p}^k_1, ..., \textbf{p}^k_{k-1}, \textbf{p}^k_{k+1},..., \textbf{p}^k_{K}),
\end{equation}
in which:
\begin{equation}
\label{equ:multinomial_parameter}
\textbf{p}^k_j=\frac{\textbf{C}_{k,j}}{ \sum_{h=\{1,2,...,k-1, k+1,...,K\}}\textbf{C}_{k,h}}.
\end{equation}
\textbf{C} is the confusion matrix given by either Eq.~(\ref{equ:normalized_cm}) or Eq.~(\ref{equ:prob_confusion_matrix}). We can see here that the classes with higher confusion are more likely to be sampled. By exposing the target class and the coupling distractors to the model, \textit{Confusion Learning} enables the model to learn better.

Combining all target classes and their corresponding distractors, we have $N_{T}^{\text{c}}$ pairs of target classes and distractors $\{(\bar{v}^{\text{c}}_1, \bar{w}^1), ..., (\bar{v}^{\text{c}}_{N_{T}^{\text{c}}}, \bar{w}^{N_{T}^{\text{c}}})\}$. For the $k$-th pair $(\bar{v}^{\text{c}}_k, \bar{w}^k)$, our learning paradigm samples $N_S$ instances for each distractor and target class, generating a support set $\mathbb{S}$ with $N_S\times (N_D+1)$ instances, and samples $N_Q$ instances of the target class $\bar{v}^{\text{c}}_k$, generating a query set $\mathbb{Q}$ with $N_Q$ instances. Combining $\mathbb{S}$ and $\mathbb{Q}$, we have a confusion task. Predictions $\text{P}(\hat{y}=k|\textbf{x}\in\mathbb{Q})$ are computed by the meta-learning algorithm using $N_{T}^{\text{c}}$ confusion tasks. By maximizing $\text{P}(\hat{y}=k|\textbf{x})$ and applying stochastic gradient descent, we can optimize parameters in the model to distinguish target classes better in the next episode until they become less confusable, after which the model switches its attention to relatively more confusable classes as the confusion matrix updates dynamically. The pseudo-code is demonstrated in Algorithm~\ref{alg:algorithm3}. 


\subsection{Confusion Matrix Estimation} 
\label{sec:3.2}
The traditional way of the confusion matrix calculation requires inferences of instances drawn from $K$ classes, which can be extremely slow and require high RAM usage when $K$ is large. To bypass this hinder, we propose a novel iterative way called Confusion Matrix Estimation (CME) to estimate the confusion matrix, the difficulty of which is much smaller than the traditional one. 

Given a trained model, we regard the confusion matrix $\textbf{C}$ calculated by evaluating the model on $K$ classes using the traditional method as the ideal confusion matrix. The purpose of CME is to estimate $\textbf{C}$ in an incremental yet less computation-consuming way. CME initializes a matrix $\textbf{E}$ of size $K \times K$, which is the same as the ideal confusion matrix. Each entry of $\textbf{E}$ is initialized with a positive constant. Since the summary of each row of $\textbf{C}$ is equal to $1$, $1/K$ is usually good for the initialization of $\textbf{E}$. The purpose of CME is to update $\textbf{E}$ in multiple steps to make it closer to the ideal confusion matrix $\textbf{C}$.  In each step, $N^\text{e}_T$ classes are uniformly sampled from meta-training dataset. Let us mark their indices among the $K$ meta-training classes as $\bar{v}^{\text{e}}=\{\bar{v}^{\text{e}}_1,...,\bar{v}^{\text{e}}_{N^\text{e}_T}\}$. By performing inference on $\bar{v}^{\text{e}}$, we are able to obtain a smaller confusion matrix named $\textbf{E}'$ with the size of  $N^\text{e}_T \times N^\text{e}_T$. Clearly, $\textbf{E}'$ contains the information on how a class is confused with other classes among $\bar{v}^{\text{e}}$. 
$\textbf{E}'$ is somehow like an observation of $\textbf{C}$ through a small ``window". To incorporate the observation into $\textbf{E}$, in each step, CME updates $\textbf{E}$ by:
\begin{equation}
\label{equ:CME_update}
\textbf{E}_{\bar{v}^{\text{e}}_i, \bar{v}^{\text{e}}_j} = \rho  \textbf{E}_{\bar{v}^{\text{e}}_i, \bar{v}^{\text{e}}_j} + (1-\rho) \textbf{E}'_{i,j}{Z},
\end{equation}
in which $\rho$ is a hyperparameter between $0$ and $1$, and $Z$ is used to scale the observation to make sure the summary of the current observation is constant with the result of previous observations:
\begin{equation}
\label{equ:CME_update_norm}
Z=\sum_{k=1,...,N^\text{e}}\textbf{E}_{\bar{v}^{\text{e}}_i, \bar{v}^{\text{e}}_k}.
\end{equation}

To combine CME with previously introduced \textit{Confusable Learning} framework, we can initialize $\textbf{E}$ and perform multi-steps CME at each \textit{Confusable Learning} episode to get a reliable estimation of the confusion matrix. However, it is not necessary to make a fresh start in each episode. Denoting the ideal confusion matrix of the model at episode $i$ as $\textbf{C}_i$, since the ability of the model will not change a lot between successive episodes, $\textbf{C}_{i+1}$ should be close to $\textbf{C}_{i}$. Intuitively, we do not need to initialize a new $\textbf{E}$ in each episode. Instead, to calculate the estimation $\textbf{E}_{i+1}$ of episode $i+1$ , we can perform CME update based on the estimation $\textbf{E}_i$ of episode $i$. As such, $\textbf{E}_0$ is initialized only once when the meta-learning model is initialized. Then at each episode, CME updates for $M$ steps. In our experience, by setting $M$ to 1, the performance of CME is good enough. Algorithm~\ref{alg:algorithm2} shows how the confusion matrix is updated in an episode. 
\begin{algorithm}[tb]
\caption{ConfusionMatrixEstimation($\mathbb{D}$, $\Theta$, $\textbf{E}$)}
\KwIn{Training set $\mathbb{D}$; Learnable weights $\Theta$; Estimation of confusion matrix $\textbf{E}$.}
\Parameter{Number of support instances for each class $N_S$; number of query instances for each class $N_Q$; number of classes to use in each CME step $N_{T}^{\text{e}}.$}
\label{alg:algorithm2}
 Initialize $\textbf{E}'$ as a zero matrix of shape $(N_{T}^{\text{e}}, N_{T}^{\text{e}})$, with each entry initialized as $0$\;
 $\bar{v}^{\text{e}} \leftarrow $\RandomSample{$\{1,...,K\}, N_{T}^{\text{e}}$}\;
 \ForEach{$k$ in $\{\bar{v}^{\text{e}}_1,...,\bar{v}^{\text{e}}_{N^\text{e}_T}\}$}{
  $\mathbb{S}_{k} \leftarrow $\RandomSample{$\mathbb{D}_{k}, N_S$}\;
  $\mathbb{Q}_{k} \leftarrow $\RandomSample{$\mathbb{D}_{k}, N_Q$}\;
 }
 
 \ForEach{$m$ in $\{1,...,N_{T}^{\text{e}}\}$}{
  \ForEach{$(\textbf{x},y)$ in $\mathbb{Q}_{\bar{v}^{\text{e}}_m}$}{
   \ForEach{$n$ in $\{1,...,N_{T}^{\text{e}}\}$}{
      Calculate $\text{P}(\hat{y}=\bar{v}^{\text{e}}_{n}|\textbf{x})$ with any base meta-learning algorithm using the support set given by $\bigcup_{k=\{\bar{v}^{\text{e}}_1,...,\bar{v}^{\text{e}}_{N^\text{e}_T}\}}\mathbb{S}_{k}$\;
      $\textbf{E}'_{m,n} \leftarrow  \textbf{E}'_{m,n}+\text{P}(\hat{y}=\bar{v}^{\text{e}}_{n}|\textbf{x})$\; \label{alg:line:cal_cm}
   }
  }
 }
 $\textbf{E}' \leftarrow \frac{\textbf{E}'}{N_Q}$\;
 Update $\textbf{E}$ using Eq.~(\ref{equ:CME_update})\;

 \Return $\textbf{E}$\;
\end{algorithm}

\begin{figure}[htb]
\subfigure[]{
\centering
\begin{minipage}[t]{1\linewidth}
\centering
\label{fig:fig_motivation:1}
\includegraphics[width=8cm]{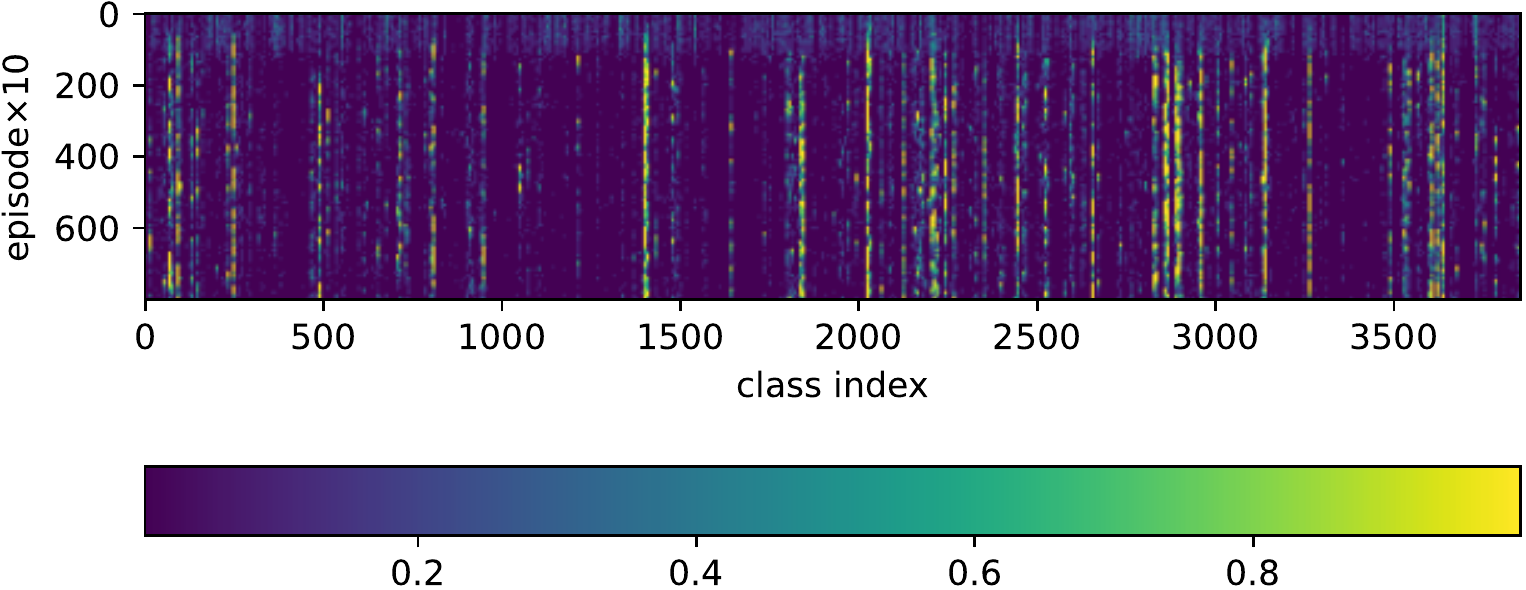} 
\end{minipage}
}
\begin{minipage}[t]{1\linewidth}
\centering
\subfigure[]{
\centering
\colorbox{green}{
\includegraphics[width=0.45cm]{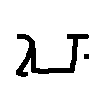}
}
\colorbox{red}{
\includegraphics[width=0.45cm]{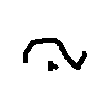}
\includegraphics[width=0.45cm]{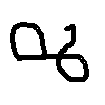}
\includegraphics[width=0.45cm]{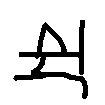}
\includegraphics[width=0.45cm]{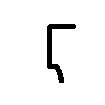}
\includegraphics[width=0.45cm]{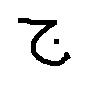}
}
\label{fig:fig_motivation:2}
}
\subfigure[]{
\centering
\colorbox{green}{
\includegraphics[width=0.45cm]{figs/fig1/0709_01.png}
}
\colorbox{red}{
\includegraphics[width=0.45cm]{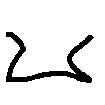}
\includegraphics[width=0.45cm]{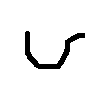}
\includegraphics[width=0.45cm]{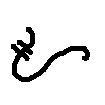}
\includegraphics[width=0.45cm]{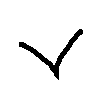}
\includegraphics[width=0.45cm]{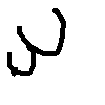}
}
\label{fig:fig_motivation:3}
}
\end{minipage}
\caption{(a) A Prototype Network is trained with \textit{Confusable Learning} on \textit{Omniglot}. Here, each pixel of the image denotes the frequency for a class to be focused, i.e., to be selected as a distractor in 10 episodes. For better visual effect, dilation has been applied to the image. (b) Target class (green) and distractors that get no attention (red) in the last 100 episodes. (c) Target class and distractors that get the most attention in the last 100 episodes.}
\label{fig:fig motivation}
\end{figure}

To demonstrate the training dynamic of our method using Algorithm \ref{alg:algorithm1}, Figure~\ref{fig:fig motivation} shows that \textit{Confusable Learning} first spreads its attention to many classes, and then turns to the classes that are more difficult to distinguish. In the early stage (top 100 rows) of meta-training, attention is dynamically spread to many classes. Later, most of these classes are well fitted by the model and get less attention. Meanwhile, other classes get more attention because they are inherently intractable to distinguish. We found that these intractable classes are visually similar (Figure ~\ref{fig:fig_motivation:3}). 

\section{Experiment}

\subsection{Experiment Setup}

\label{experiment setup}

To empirically prove the efficacy of our method, we conduct experiments on three real-world datasets: \textit{Omniglot}, \textit{Fungi}, and \textit{ImageNet}. We choose these datasets because they contain more than 1000 classes, unlike prior works on meta-learning which experiment with smaller images and fewer classes.

\textit{Confusable Learning} can be easily applied to various mainstream meta-learning algorithms. To demonstrate the efficacy of our method, we choose four state-of-the-art meta-learning algorithms as our base meta-learning methods in our experiments: Prototypical Network (PN)~\cite{snell2017prototypical}, Matching Network (MN)~\cite{vinyals2016matching}, Prototype Matching Network (PMN)~\cite{wang2018low} and Ridge Regression Differentiable Discriminator (R2D2)~\cite{bertinetto2018metalearning}. We will show that by applying \textit{Confusable Learning} to these methods, we can easily improve their performance in large-class few-shot learning setting. For notation, we denote our method by attaching a ``w/CL" behind each of them. For example, PN w/CL means prototype networks with \textit{Confusable Learning}.

In few-shot learning, $N$-shot $K$-way classification tasks consist of $N$ labeled instances for each $K$ classes. As stated in PN~\cite{snell2017prototypical}, it can be extremely beneficial to train meta-learning models with a higher way than that will be used in meta-testing dataset. Particularly, for PN, to train a model for 5/20 ways tasks, the author used training tasks with 60 classes~\cite{snell2017prototypical}. The same setting is also mentioned in other mainstream few-shot methods ~\cite{bertinetto2018metalearning}. However, for the large-class few-shot problem which has a large number of classes, it becomes impractical to build even larger support sets due to the limitation of memory and the exponentially growing load of computation. Therefore, in our experiment, the models are all trained in meta-training tasks with fewer ways than the meta-testing task. We evaluate the models using query sets and support sets constructed in the same way as their authors originally did ~\cite{vinyals2016matching,snell2017prototypical,wang2018low,bertinetto2018metalearning}.



\subsubsection{\textit{Omniglot}.}

\textit{Omniglot} contains 1623 handwritten characters. As what Vinyals~\cite{vinyals2016matching} has done, we resize the image to $28\times28$ and augment the dataset by rotating each image by 90, 180, and 270 degrees. For a more challenging meta-testing environment, we employ the split introduced by Lake~\cite{lake2015human}, constructing our meta-training dataset with 3856 classes and meta-testing dataset with 2636 classes. In meta-testing stage, we use all 2636 classes in every single meta-testing task.  

For PN, MN and PMN, learning rate is set to 1e-4. For R2D2, learning rate is set to 5e-5. For our method, We set $N_D$ to $40$, $N_{T}^{\text{e}}$ to 500. In our experiments of all datasets, $N_{T}^{\text{c}}$ is always set to $(N^{\text{e}}_T \times 2) / (N_D + 2)$, which is 23 here. $\rho$ is set to $0.9$. $M$ is set to $1$.


\begin{figure}[!ht]
\centering
\subfigure[PN vs. PN w/CL]{
\centering
\includegraphics[width=2.7cm]{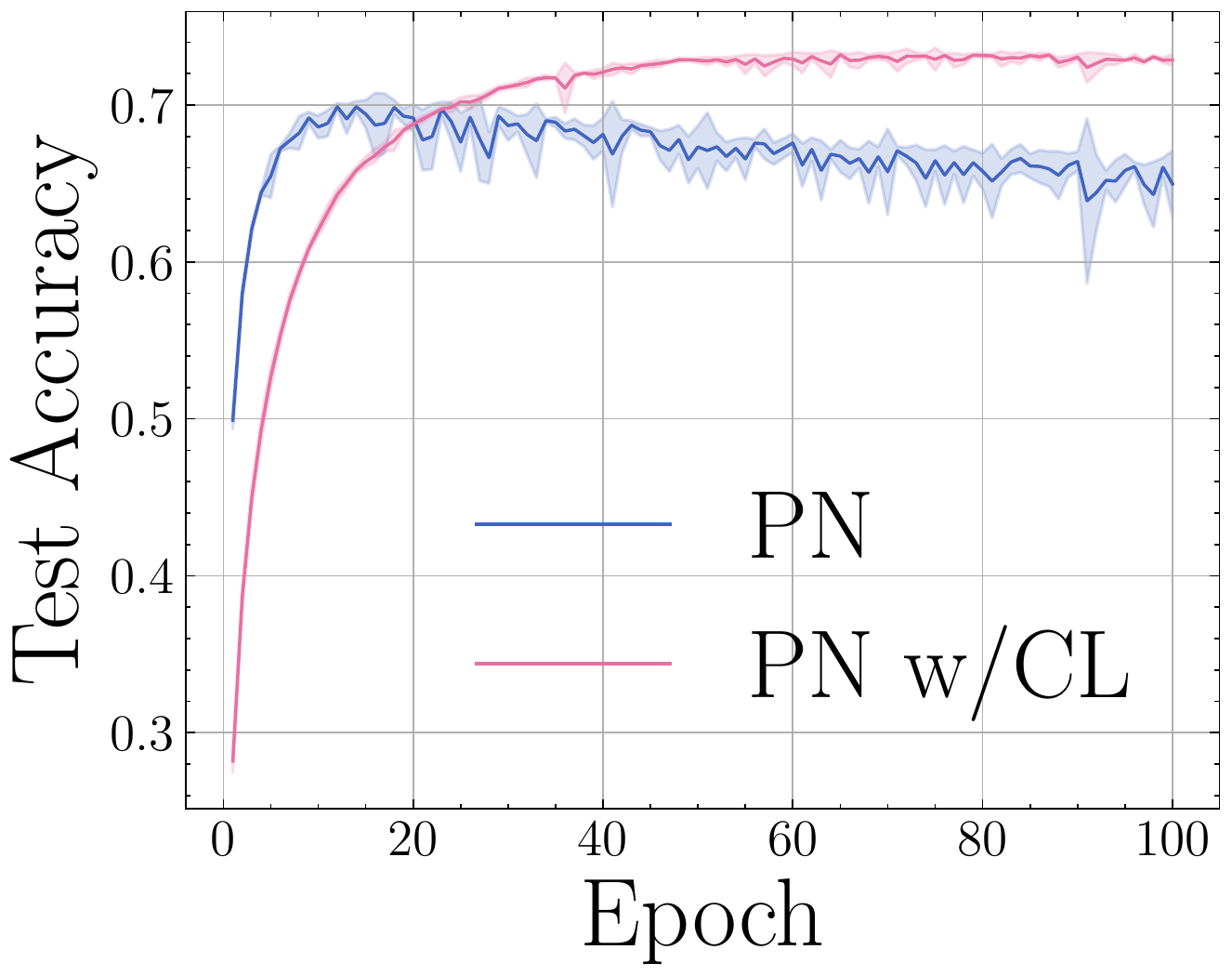}
\label{fig:omniglot:pn}
}
\subfigure[MN vs. MN w/CL]{
\centering
\includegraphics[width=2.7cm]{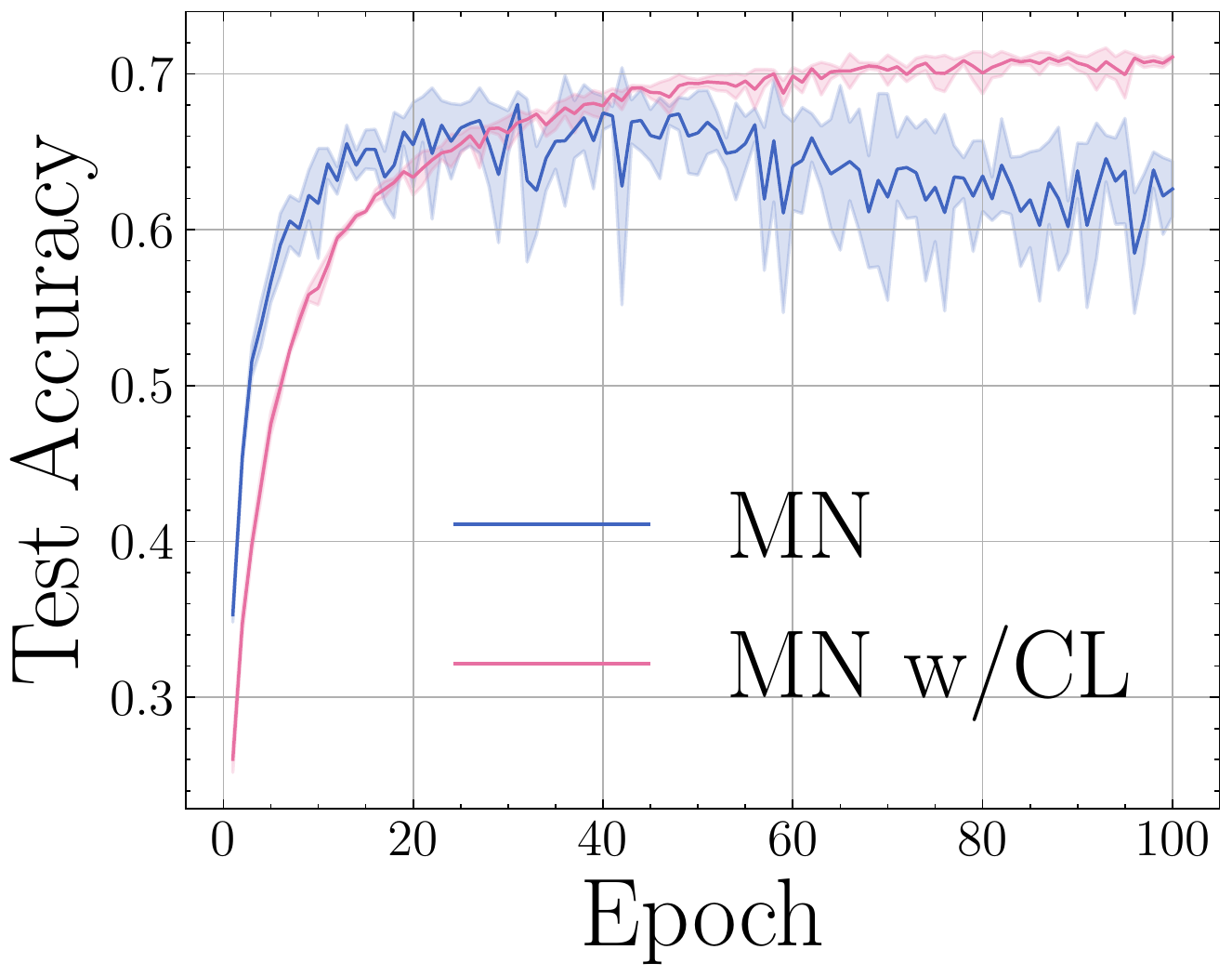}
\label{fig:side:b}
}
\subfigure[PMN vs. PMN w/CL]{
\centering
\includegraphics[width=2.7cm]{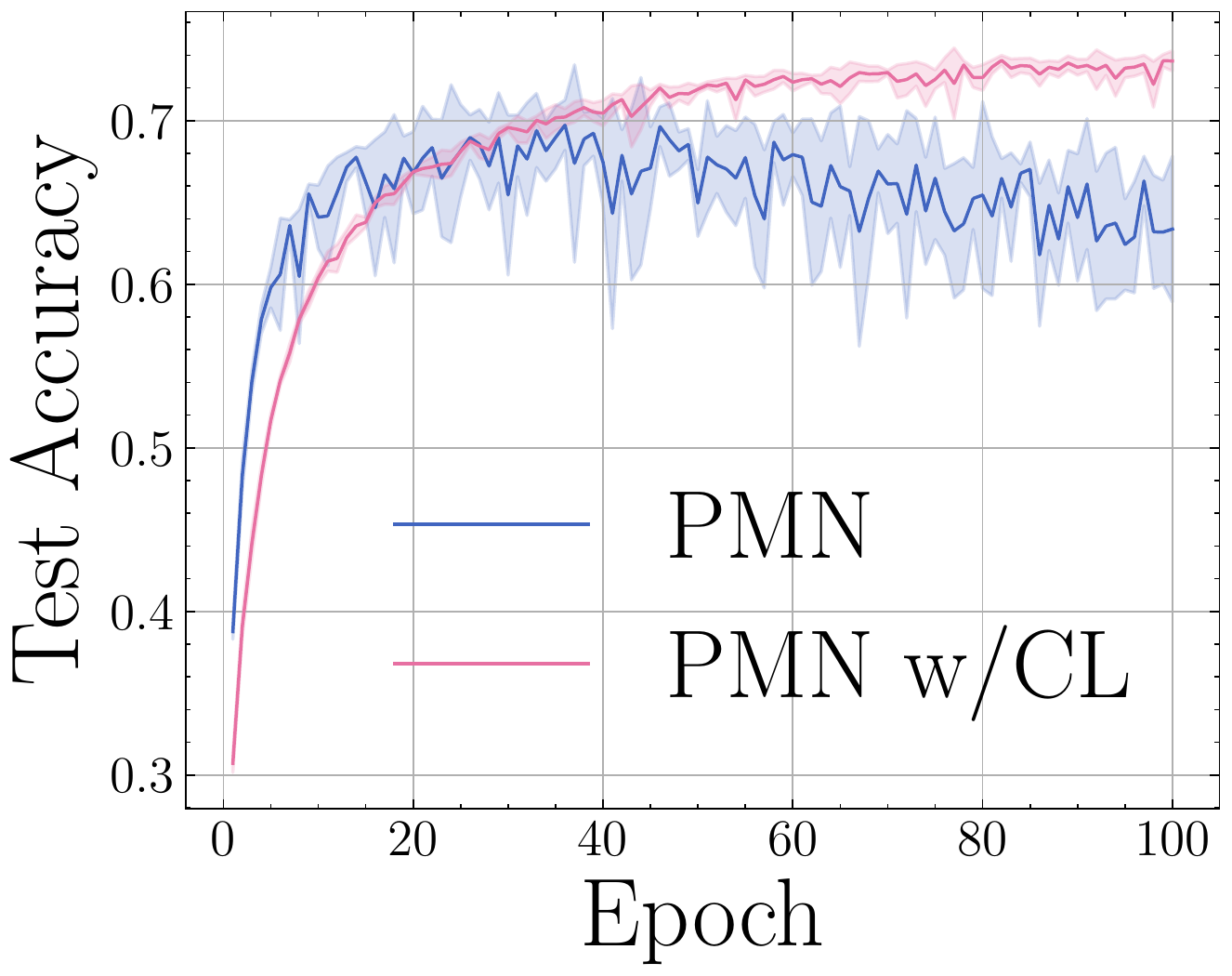}
\label{fig:side:c}
}
\subfigure[R2D2 vs. R2D2 w/CL]{
\centering
\includegraphics[width=2.7cm]{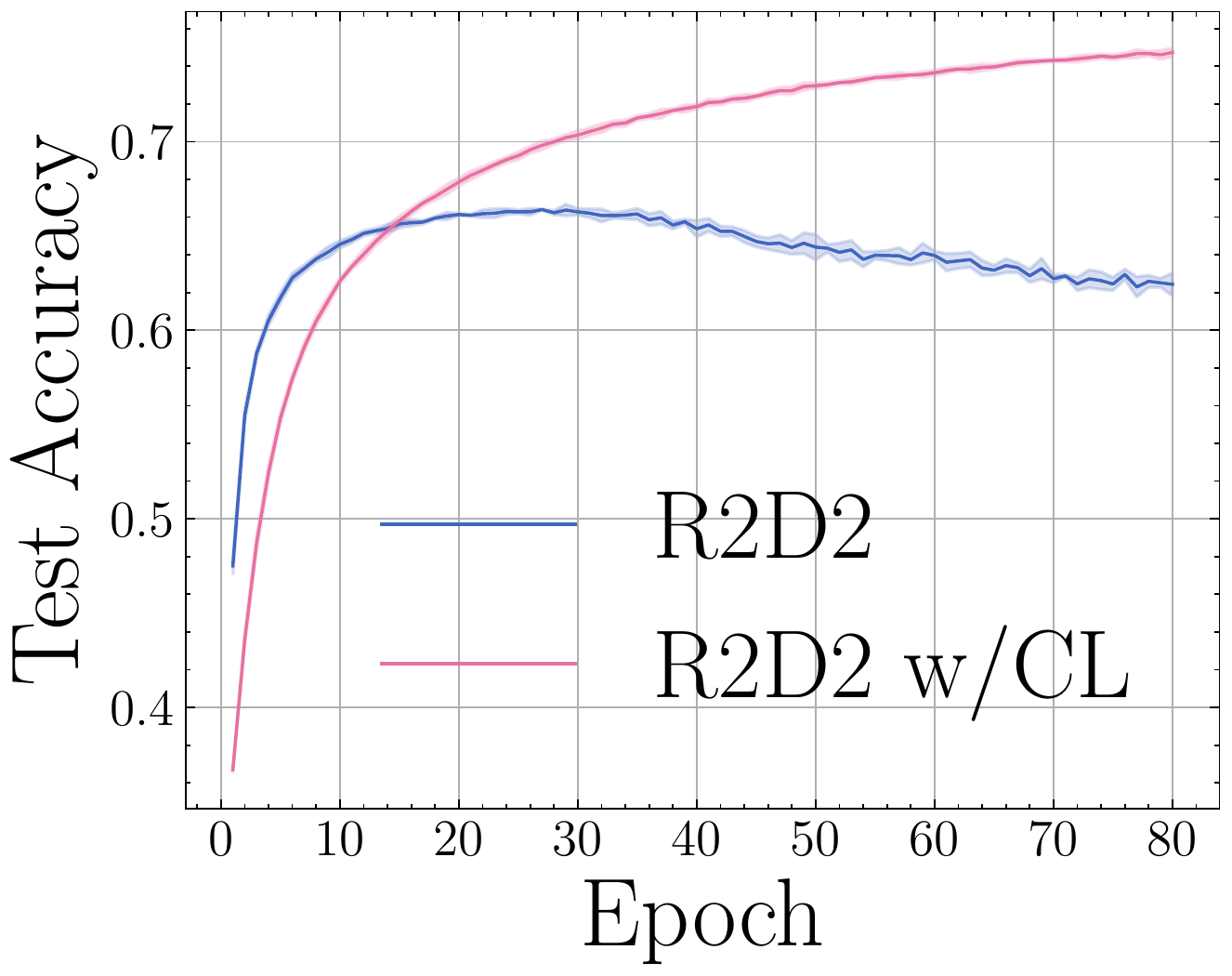}
\label{fig:side:c}
}
\caption{Test accuracy vs. number of epoch of 4 kinds of meta-learning method without vs. with \textit{Confusable Learning} in \textit{Omniglot}. }
\label{fig:fig1}
\end{figure}

\begin{figure}[!ht]
\centering
\subfigure[PN vs. PN w/CL]{
\centering
\includegraphics[width=2.7cm]{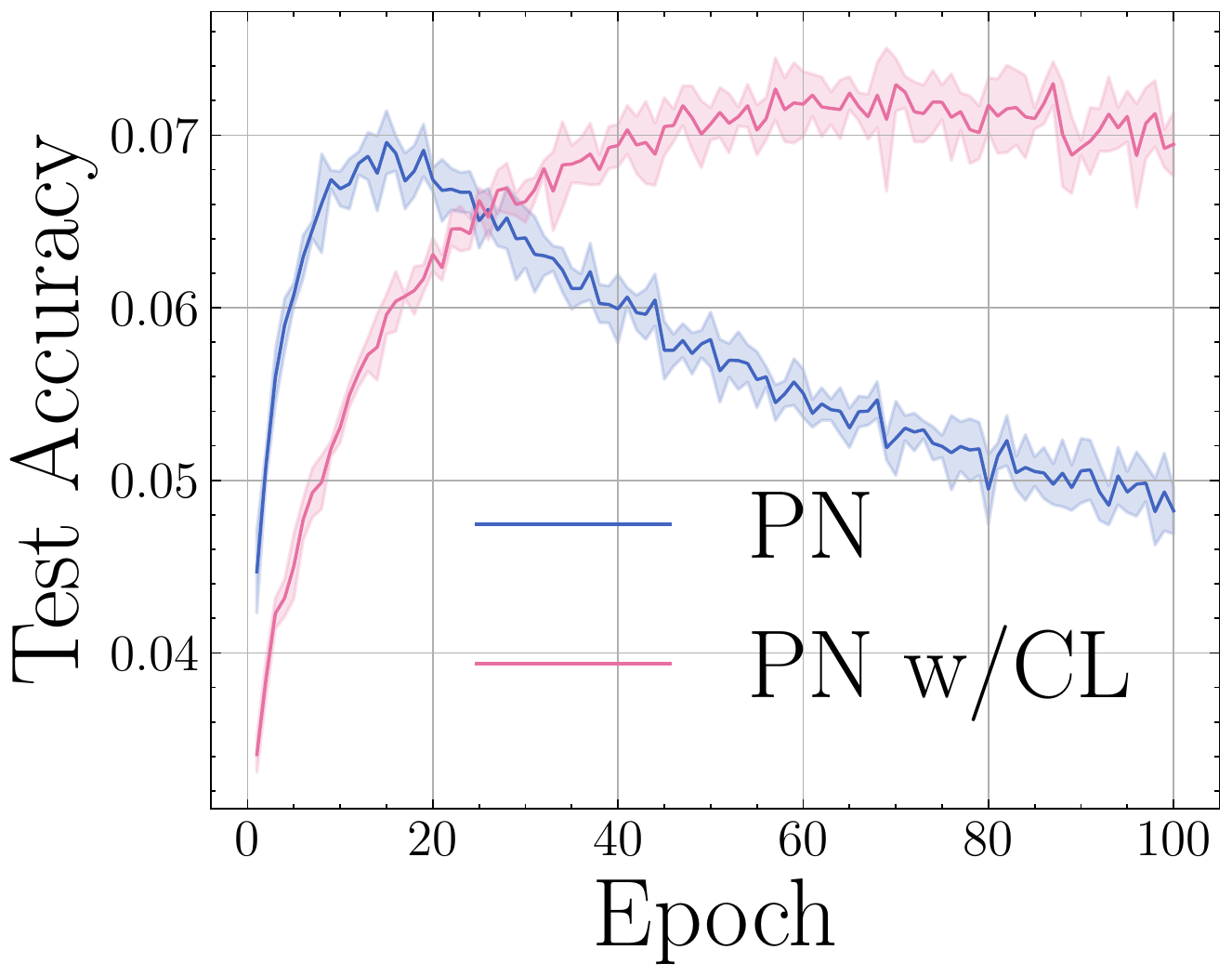}
\label{fig:side:a}
}
\subfigure[MN vs. MN w/CL]{
\centering
\includegraphics[width=2.7cm]{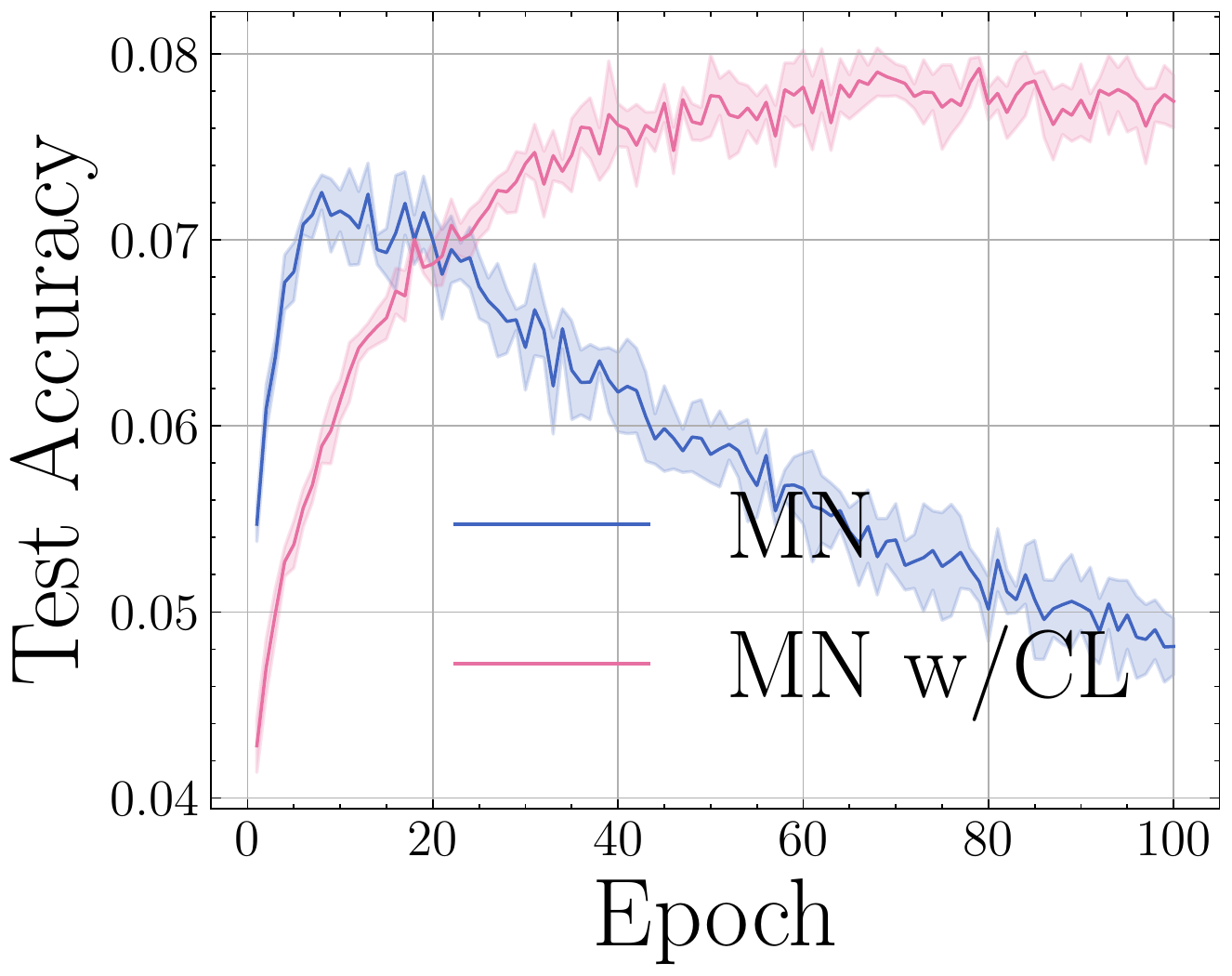}
\label{fig:side:b}
}
\subfigure[PMN vs. PMN w/CL]{
\centering
\includegraphics[width=2.7cm]{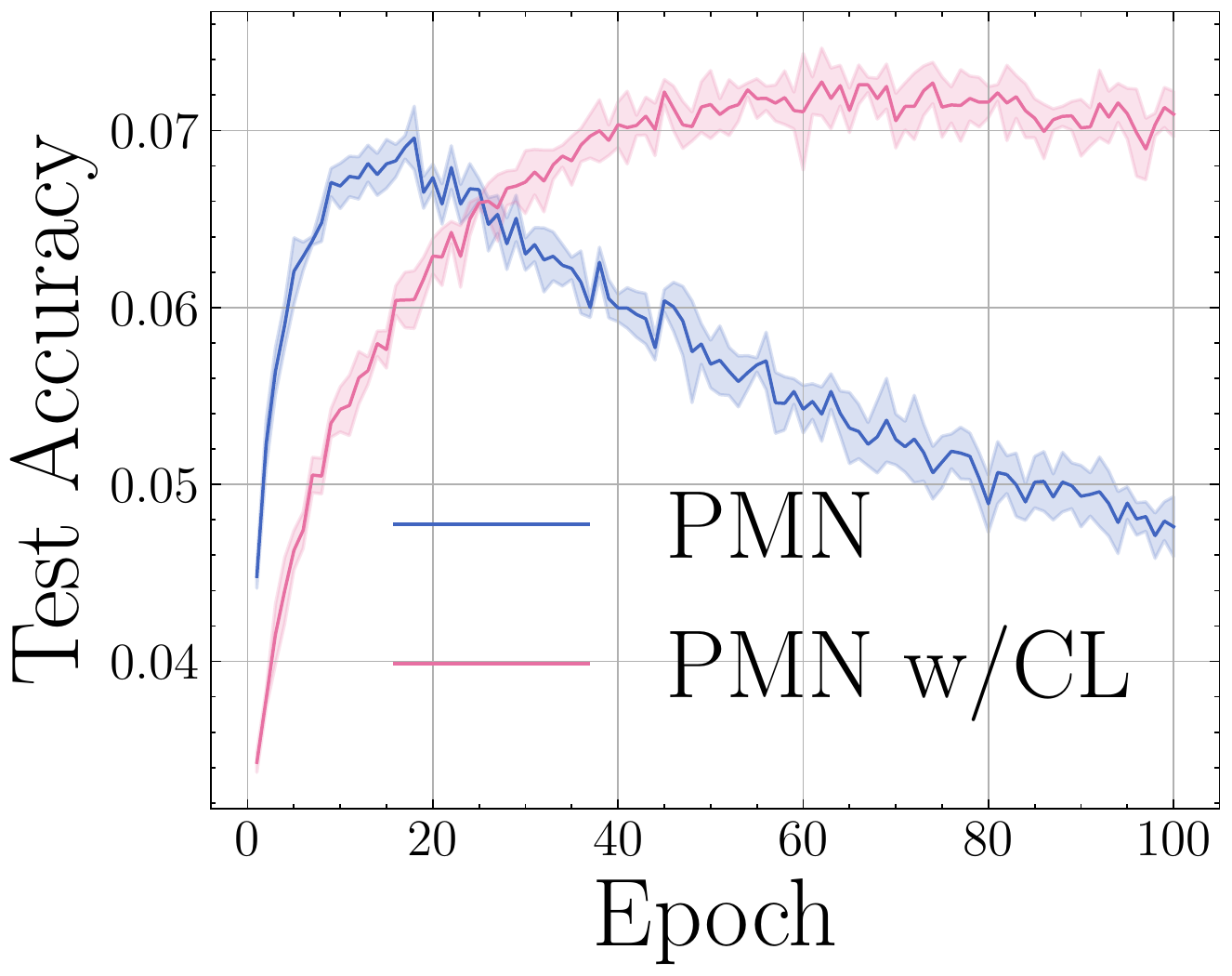}
\label{fig:side:c}
}
\subfigure[R2D2 vs. R2D2 w/CL]{
\centering
\includegraphics[width=2.7cm]{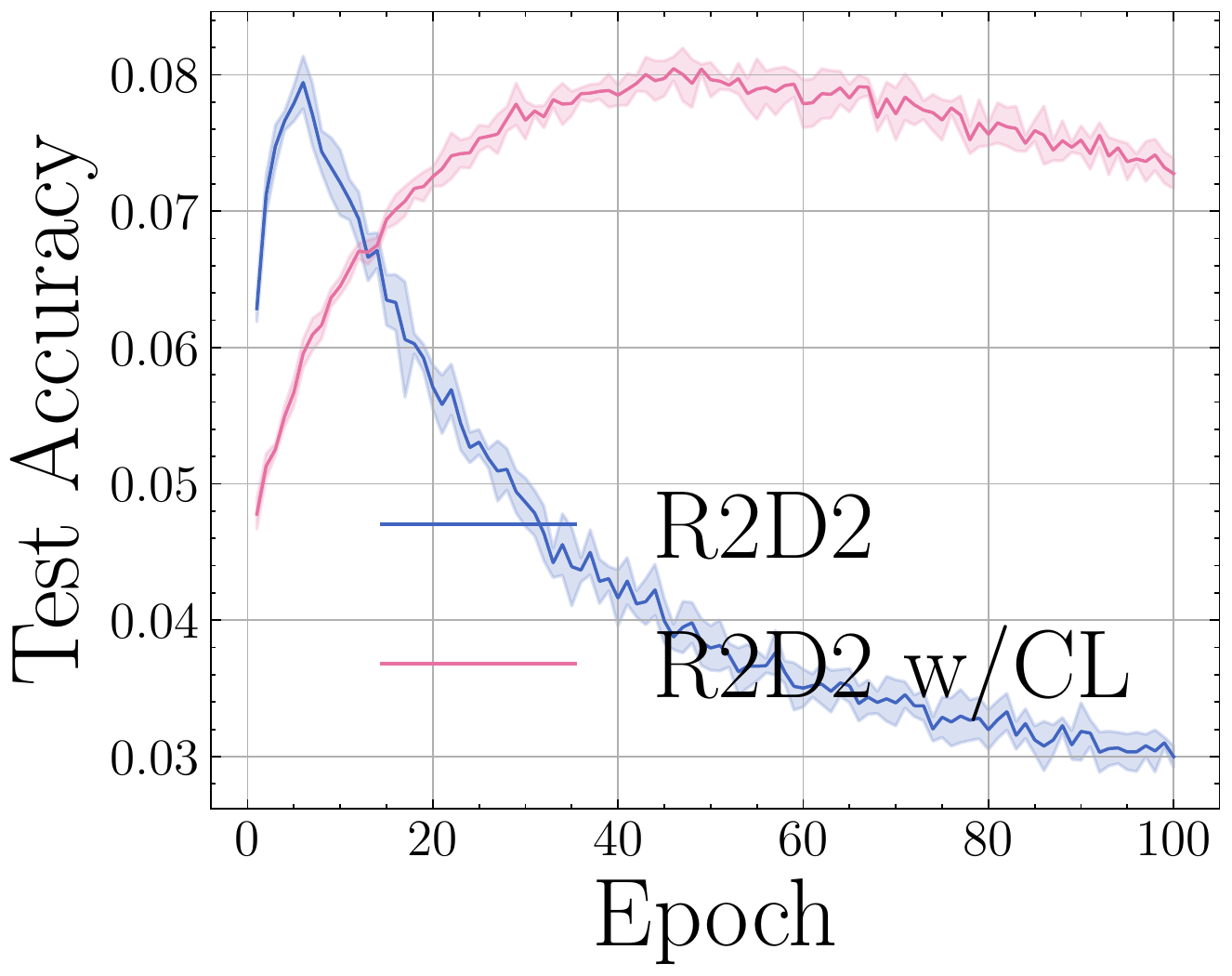}
\label{fig:side:c}
}
\caption{Test accuracy vs. number of epoch of 4 kinds of meta-learning method without vs. with \textit{Confusable Learning} in \textit{fungi}.}
\label{fig:figfffff}
\end{figure} 

\begin{figure}[!ht]
\centering
\subfigure[PN vs. PN w/CL]{
\centering
\includegraphics[width=2.7cm]{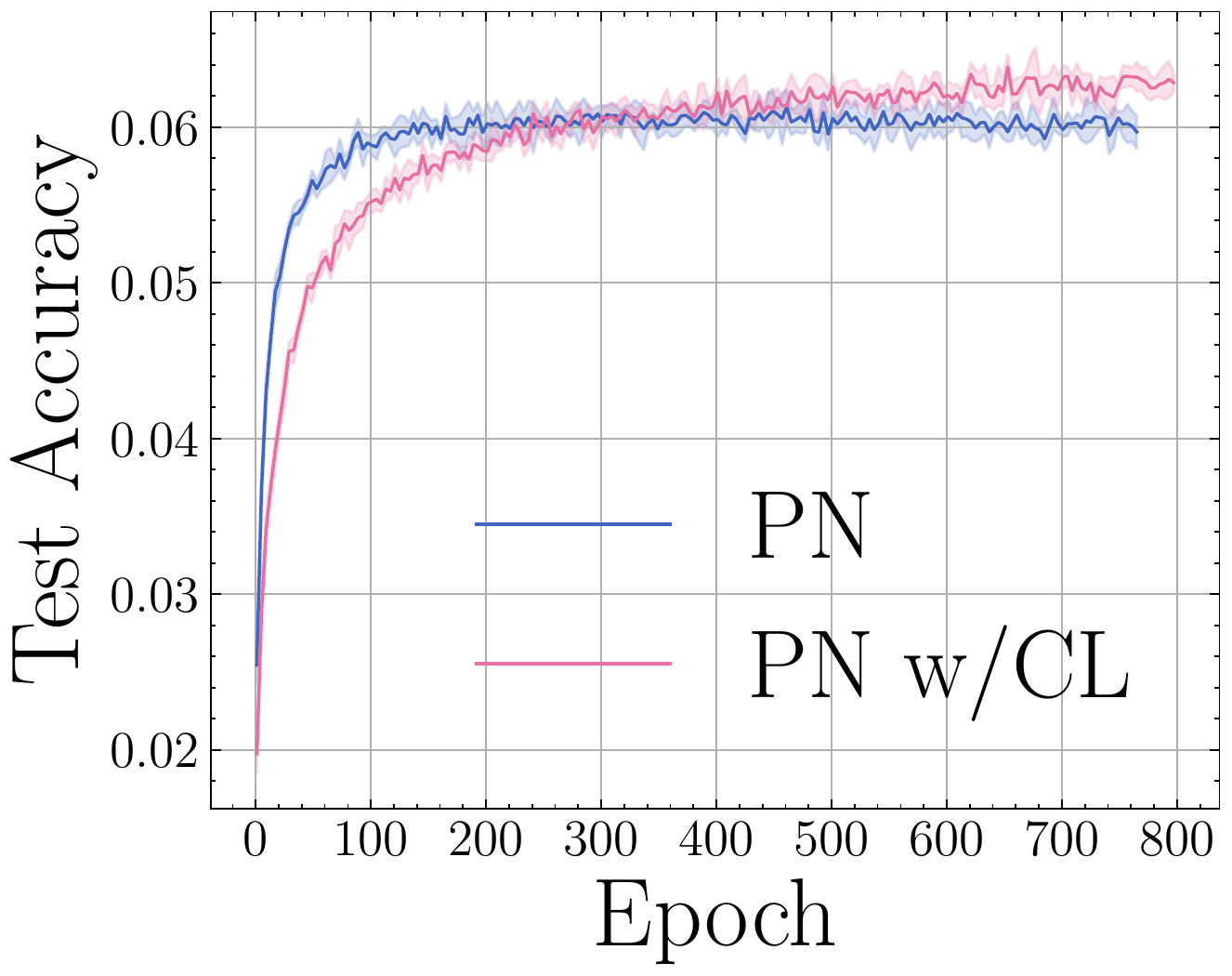}
\label{fig:side:a}
}
\subfigure[MN vs. MN w/CL]{
\centering
\includegraphics[width=2.7cm]{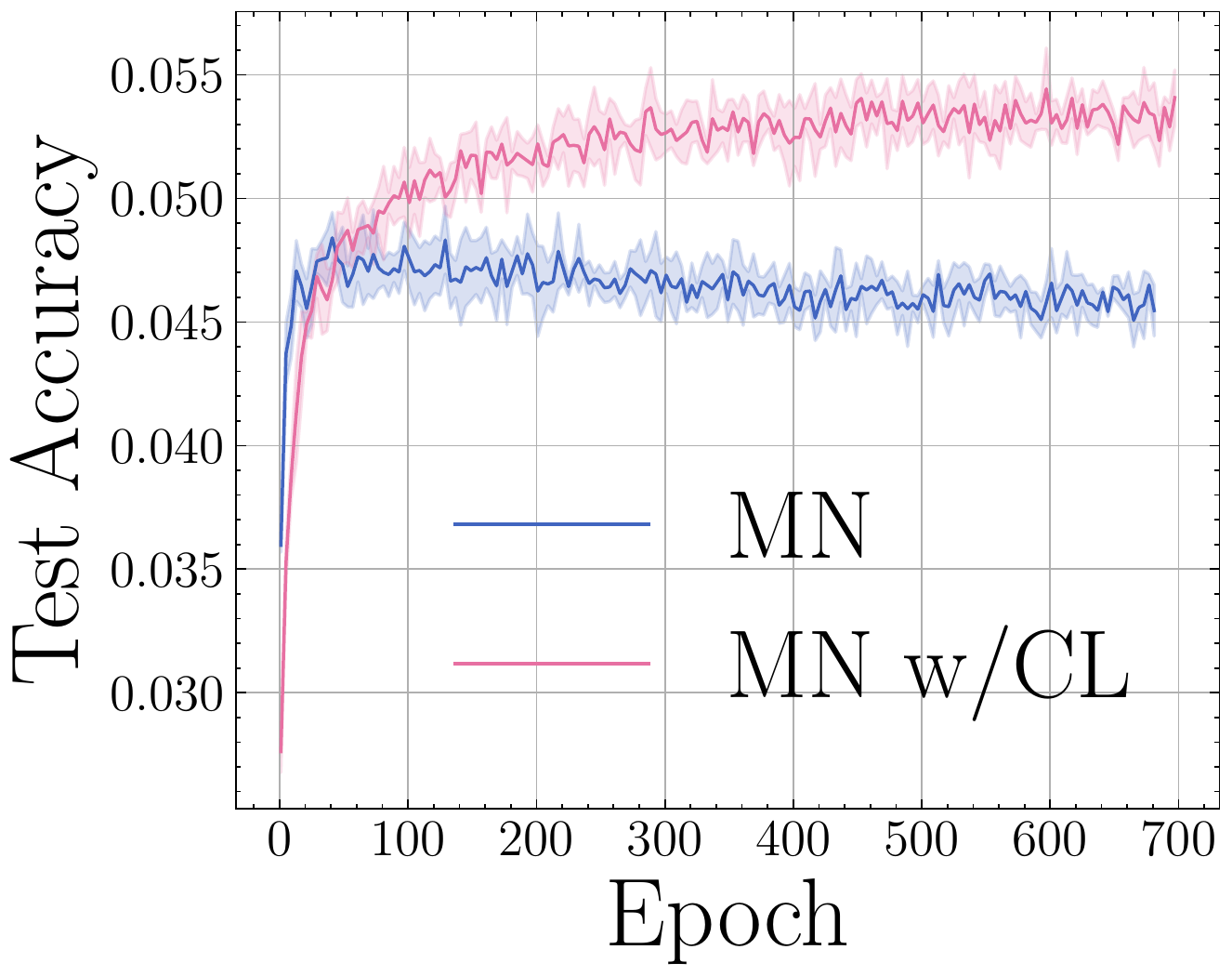}
\label{fig:side:b}
}
\subfigure[PMN vs. PMN w/CL]{
\centering
\includegraphics[width=2.7cm]{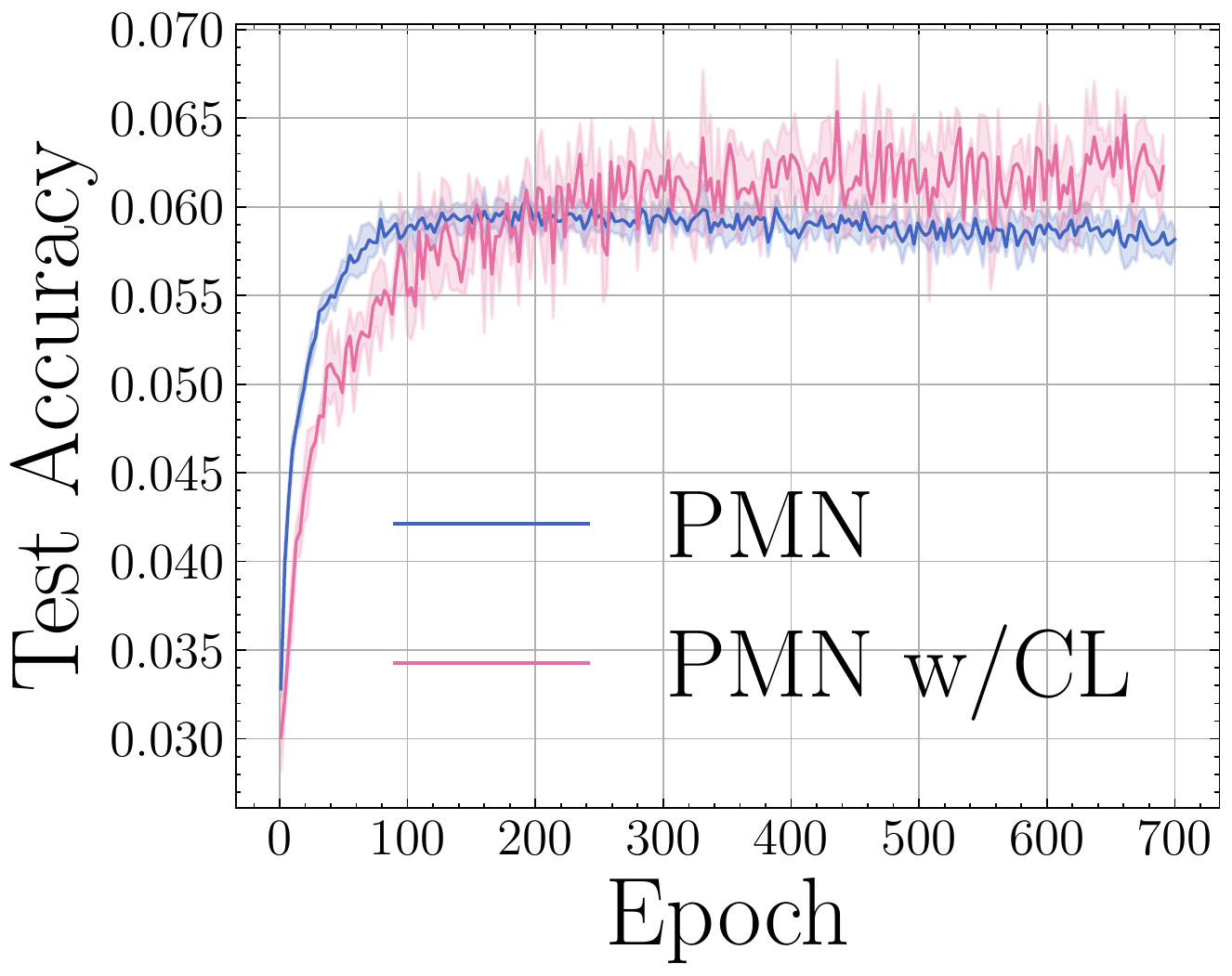}
\label{fig:side:c}
}
\subfigure[R2D2 vs. R2D2 w/CL]{
\centering
\includegraphics[width=2.7cm]{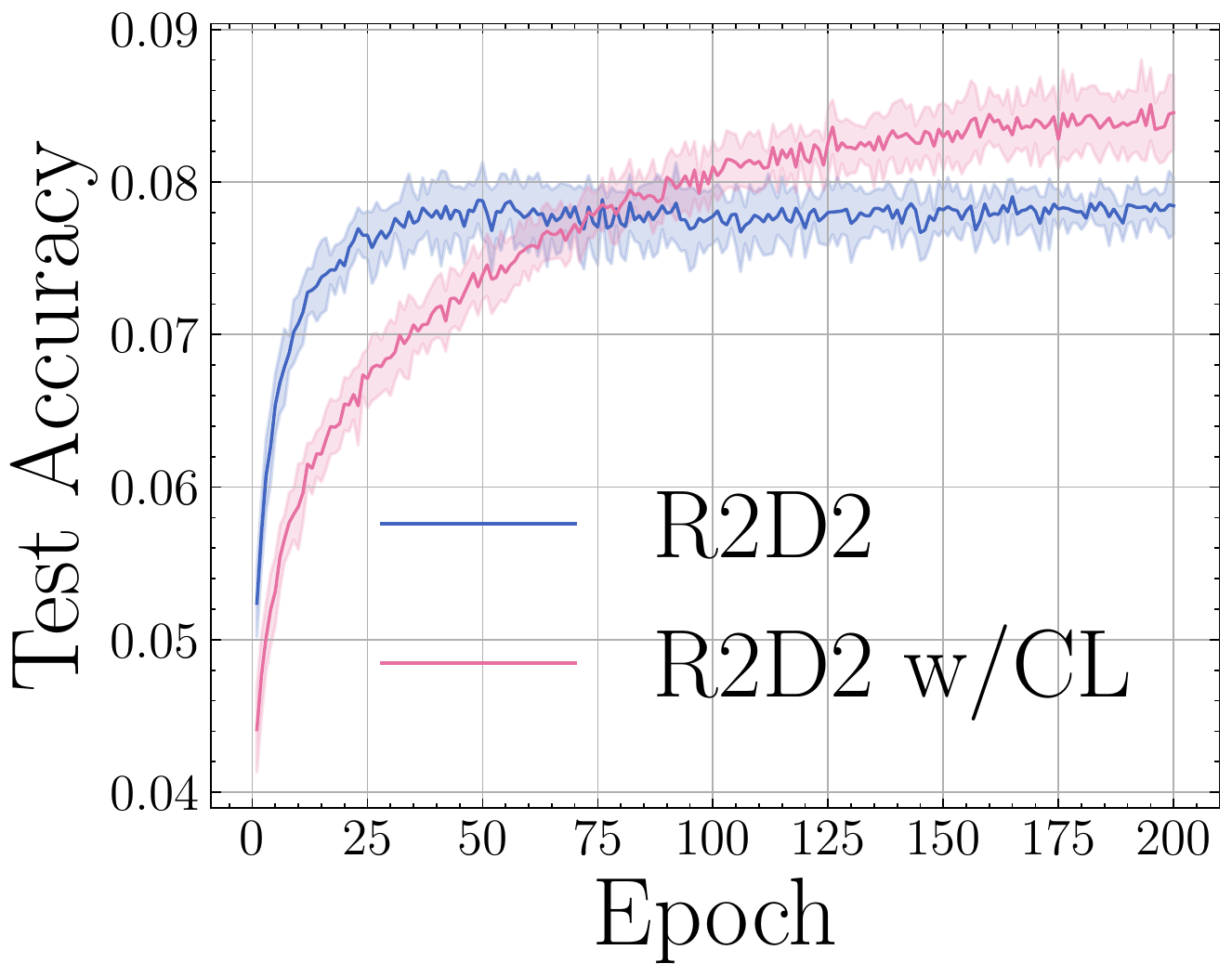}
\label{fig:side:ca}
}
\caption{Test accuracy vs. number of epoch of 4 kinds of meta-learning methods without vs. with \textit{Confusable Learning} in \textit{Imagenet}.}
\label{fig:fig10}
\end{figure}

\subsubsection{\textit{Fungi}.}
\textit{Fungi} is originally introduced by the 2018 FGVCx Fungi Classification Challenge. We randomly sample 632 classes to construct meta-training dataset, 674 classes to construct meta-testing dataset and 88 classes to construct validation dataset. All 674 meta-testing classes are used in every meta-testing task. 

Learning rate is set to 5e-4 for all methods. For our method, We set $N_D$ to $5$. $N_{T}^{\text{e}}$ is set to $70$, and thus $N_{T}^{\text{c}}$ is set to $20$. $\rho$ is set to $0.9$. $M$ is set to $1$. For R2D2 w/CL, the temperature of the softmax is set to $0.1$ in CME.

\subsubsection{\textit{ImageNet64x64}.}
We also conduct experiments on \textit{ImageNet64x64} dataset ~\cite{chrabaszcz2017downsampled}, which is a downsampled version of the original \textit{ImageNet} used in ILSVRC with images resized to 64$\times$64 pixels. The reason why we do not perform experiments on regular few-shot datasets like \emph{mini}ImageNet and \emph{tiered}ImageNet is that neither of them holds enough number of classes for our large-class setting. Although \textit{ImageNet64x64} is not regularly used to evaluate few-shot learning methods, it contains 1000 classes in total. Here we merge its original training dataset and original testing dataset together and then split it by class again into a new meta-training dataset, a validation dataset, and a new meta-testing dataset according to the category of the classes. All classes that belong to the category \textbf{Living Thing} are used for meta-training stage and all classes that belong to the category \textbf{Artifact} are used for meta-testing stage. This gives us a meta-training dataset with 522 classes and a meta-testing dataset with 410 classes. The rest 68 classes belong to the validation dataset.

For MN and MN w/CL, we set the learning rate to 1e-3. For the other settings, the learning rate is set to 1e-4. In our experiment, We set $N_D$ to $5$. $N_{T}^{\text{e}}$ is set to $90$, and thus $N_{T}^{\text{c}}$ is set to $25$. $\rho$ is set to $0.9$. $M$ is set to $1$.

\subsection{Result}





We perform five repeated experiments with different random seeds. In Table~\ref{tab:acc}, the averaged accuracy for each method is shown. It can be seen all meta learners coupled with our method outperform original ones in all three datasets. 



To delve into the training dynamic, Figures~\ref{fig:fig1},~\ref{fig:figfffff} and~\ref{fig:fig10} show how testing accuracy changes with respect to the number of epochs, with the standard deviation shown with the shaded area. The meta-learning algorithm with \textit{Confusable Learning} has a better generation than those without it.  
As shown in Figure~\ref{fig:figfffff}, it is interesting to find out that \textit{Confusable Learning} shows a great ability to resist over-fitting than corresponding original methods in \textit{Fungi}, which contains very similar mushroom classes and thus leads to over-fitting easily.

\begin{table}[!htb]
\centering
\scalebox{1}{
\begin{tabular}{|l|r|r||r|r||r|r||r|r|}  
\hline
Algorithm & PN & \begin{tabular}{@{}c@{}}PN \\ w/CL\end{tabular}  & MN & \begin{tabular}{@{}c@{}}MN \\ w/CL\end{tabular}  & PMN & \begin{tabular}{@{}c@{}}PMN \\ w/CL\end{tabular} & R2D2 & 
\begin{tabular}{@{}c@{}}R2D2 \\ w/CL\end{tabular}  \\
\hline

\textit{Omniglot} & 69.90\% & \textbf{73.21\%} & 68.04\%  & \textbf{71.07\%} & 69.73\% & \textbf{73.68\%} & 66.12\% & \textbf{74.72\%} \\
\hline

\textit{Fungi} & 6.96\% & \textbf{7.29}\% & 7.26\% & \textbf{7.92\%} & 6.96\% & \textbf{7.28\%} & 7.92\% & \textbf{8.04\%} \\
\hline

\textit{ImageNet} &  6.02\% & \textbf{6.27\%} & 4.79\% & \textbf{5.41\%} & 5.90\%  & \textbf{6.41\%} & 7.78\%  & \textbf{8.46\%}\\
\hline

\end{tabular}
}
\caption{5-shot classification test accuracies of PN, PN w/CL, MN, MNw/CL, PMN, PMNw/CL and R2D2, R2D2 w/CL in \textit{Omniglot} (2636-way), \textit{Fungi} (674-way) and \textit{ImageNet64x64} (410-way).}
\label{tab:acc}
\end{table}

\begin{figure}[!htb]
\centering
\subfigure[\textit{Omniglot}]{
\centering
\includegraphics[width=2.7cm]{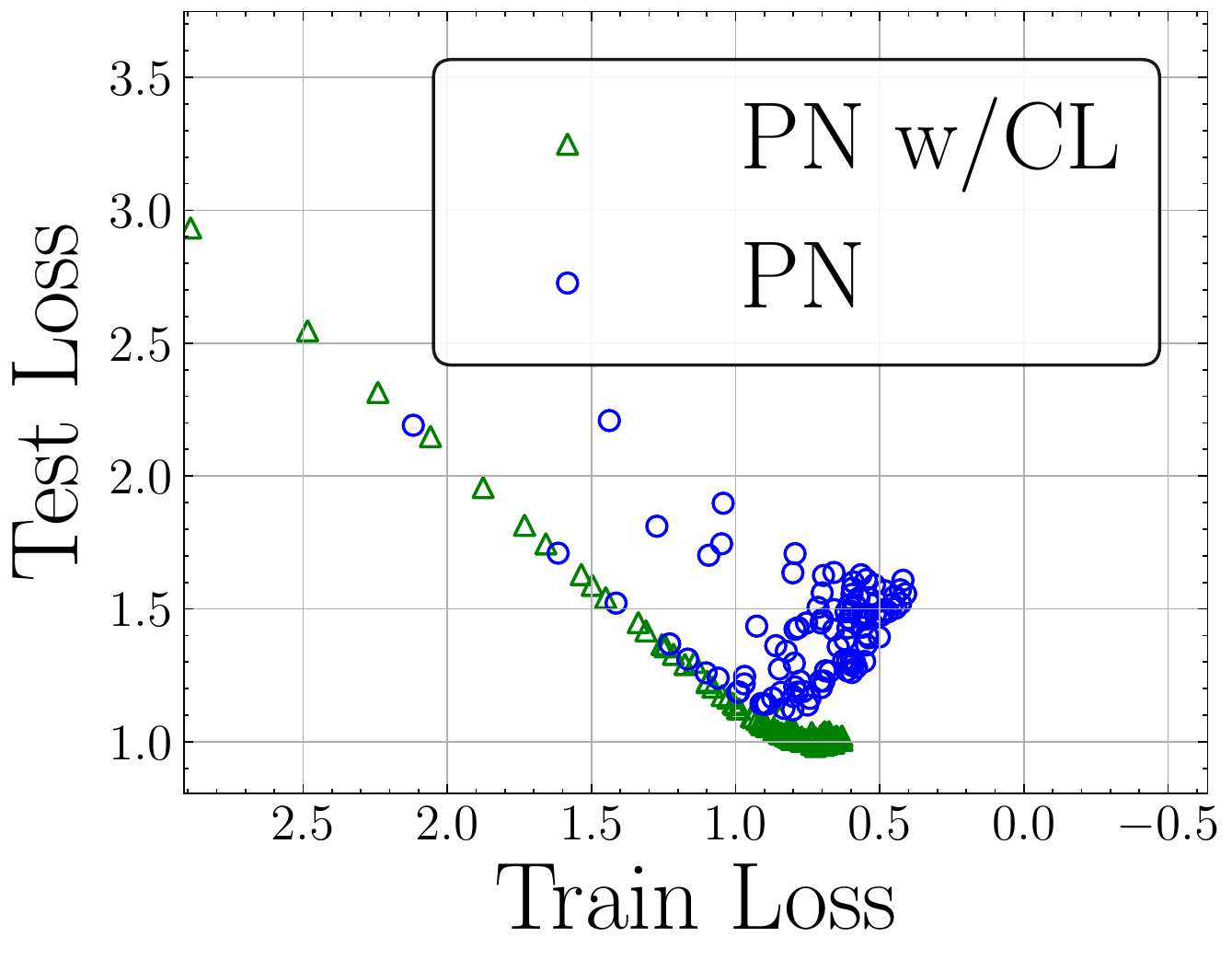}
\label{fig:generalization:a}
}
\subfigure[\textit{Fungi}]{
\centering
\includegraphics[width=2.7cm]{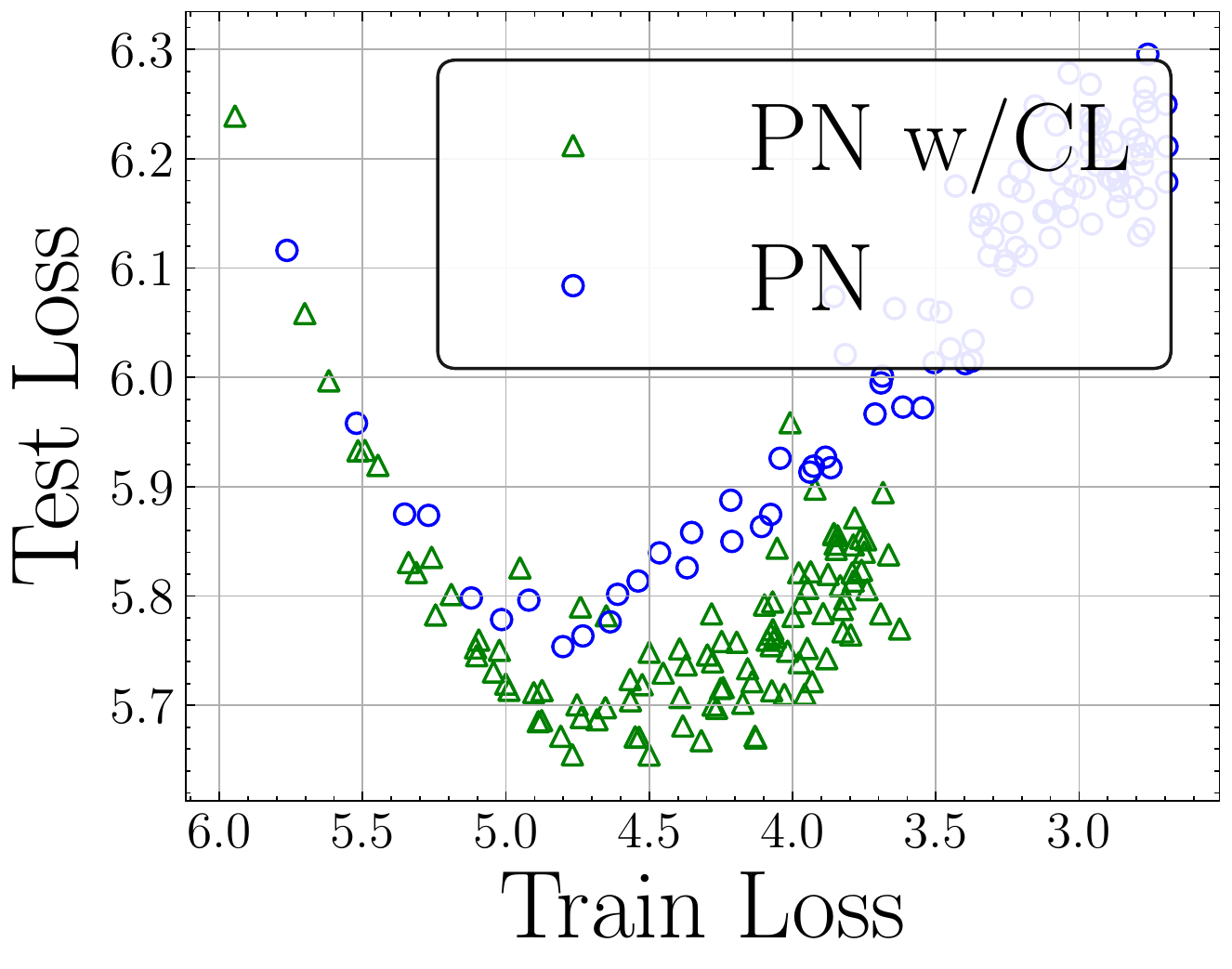}
\label{fig:generalization:b}
}
\subfigure[\textit{ImageNet64x64}]{
\centering
\includegraphics[width=2.7cm]{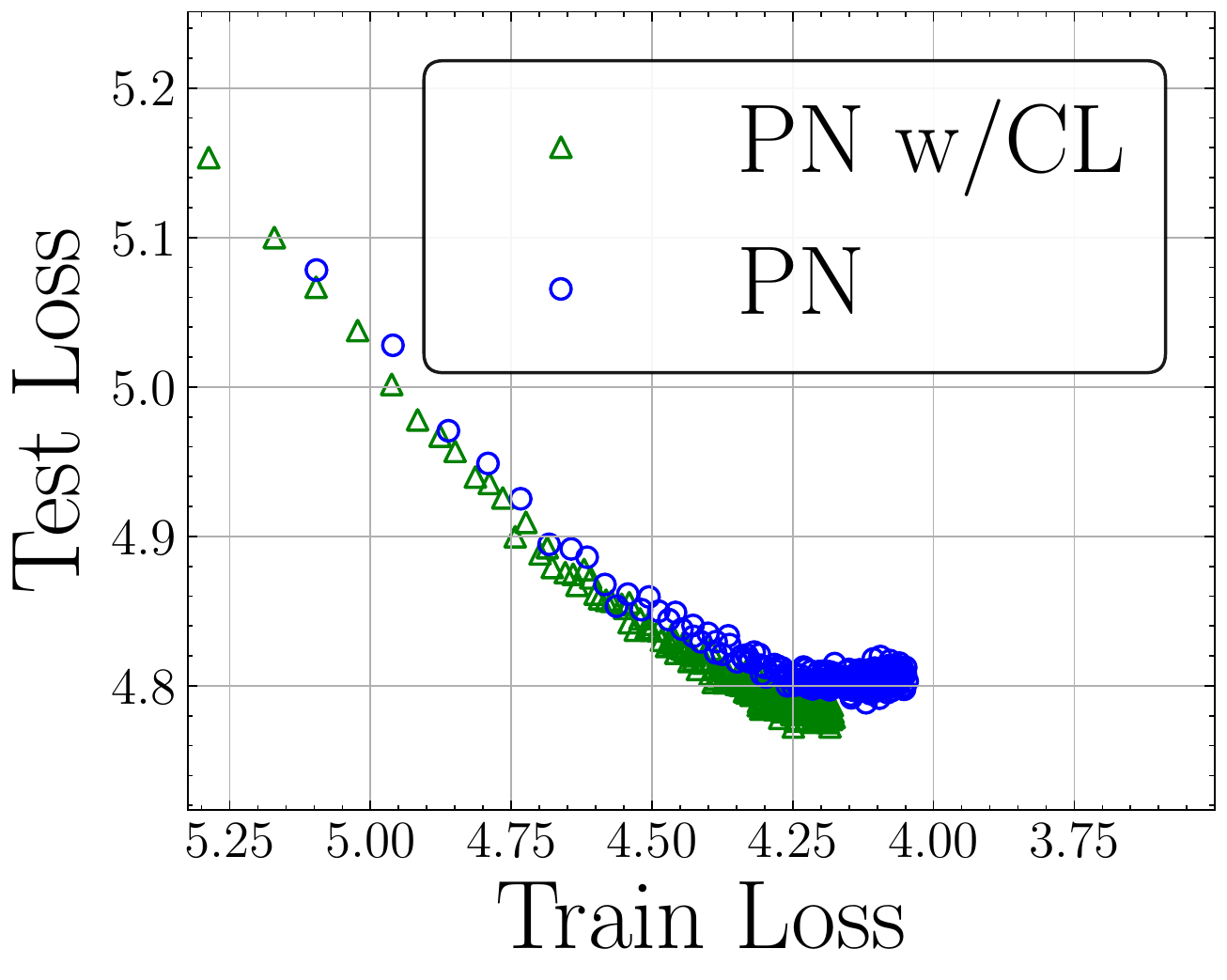}
\label{fig:generalization:c}
}

\caption{(a) Training loss vs. testing loss in \textit{Omniglot} dataset. The triangles or circles on the right have smaller training loss, and thus usually represent the performance of the meta-learning algorithm in the late stage of training (b) Training loss vs. testing loss in \textit{Fungi} dataset. (b) Training loss vs. testing loss in \textit{ImageNet64x64} dataset.}
\label{fig:generalization}

\end{figure}

To further demonstrate the ability of \textit{Confusable Learning} to resist over-fitting, we visualize the co-relationship between the training loss and the testing loss of PN and PN w/CL. Note that we estimate the training loss of the model using query sets and support sets constructed in the same way as Snell~\cite{snell2017prototypical}, but including all classes in the meta-training dataset in each task. To be specific, we estimate training losses in \textit{Omniglot} (3856-way 5-shot), \textit{Fungi} (632-way 5-shot), and \textit{ImageNet64x64} (522-way 5-shot). Figure~\ref{fig:generalization} shows that under the same training loss, \textit{Confusable Learning} has smaller testing loss. We believe that PN over-fits the regular patterns and ignores confusable patterns, while \textit{Confusable Learning} focuses on these confusable patterns. 

\subsection{Ablation Study}

In this section, we will discuss the influence of proposed soft confusion matrix and Confusion Matrix Estimation in~\textit{Confusable Learning}.

\subsubsection{Influence of calculating confusion matrix with probability.}

\begin{table}[!]

    \centering
    
    \begin{tabular}{|l|r|r|r|r|}
    \hline
        Algorithm & PN & MN & PMN & R2D2 \\
    \hline
        baseline & 69.90$\pm$0.16\% & 68.04$\pm$0.23\% & 69.73$\pm$0.21\% & 66.12$\pm$0.16\%  \\
    \hline
        w/CLN & \textbf{73.23$\pm$0.10\%} & 70.63$\pm$0.09\% & 73.24$\pm$0.09\% & 74.65$\pm$0.14\% \\
    \hline
        w/CL & 73.21$\pm$0.04\% & \textbf{71.07$\pm$0.19\%} & \textbf{73.68$\pm$0.21\%} &  \textbf{74.72$\pm$0.52\%} \\
    \hline
    \end{tabular}
    
    \caption{5-shot classification test accuracies on \textit{Omniglot} for \textit{Confusable Learning} using the traditional confusion matrix and the proposed soft confusion matrix.}
    \label{table:ablation_count_omniglot}
\end{table}{}

\textit{Confusable Learning} adopts a novel definition of confusion matrix called soft confusion matrix as shown in Eq.~(\ref{equ:prob_confusion_matrix}). To demonstrate its benefit, we implement \textit{Confusable Learning} using the traditional definition of confusion matrix given in Eq.~(\ref{equ:normalized_cm}). To implement such a setting, we simply replace $\text{P}(\hat{y}=\bar{v}^{\text{e}}_{n}|\textbf{x})$ in line \ref{alg:line:cal_cm} of Algorithm~\ref{alg:algorithm2} with $\textbf{1}_{\bar{v}^{\text{e}}_{n}=\mathop{\arg\max}_{k}P(\hat{y} = k | (\textbf{x}, y))}$. We denote this setting by attaching a ``w/CLN" behind the name of each meta-learning algorithm. Results of \textit{Omniglot} and \textit{Fungi} are shown in Table \ref{table:ablation_count_omniglot} and Table \ref{table:ablation_count_fungi} respectively. It can be seen that the accuracy of the model using soft confusion matrix, marked with ``w/CL", is higher than the model marked as ``w/CLN" on \textit{Fungi}. However, on \textit{Omniglot} they are almost equal. This is because the model is more confident about its prediction on \textit{Omniglot}, making $P(\hat{y}|(\textbf{x}, y))$ close to either $0$ or $1$. In this case, the proposed soft confusion matrix is equivalent to the traditional one.

\begin{table}[!]

    \centering
    
    \begin{tabular}{|l|r|r|r|r|}
    \hline
        Algorithm & PN & MN & PMN & R2D2  \\
    \hline
        baseline & 6.96$\pm$0.07\% & 7.26$\pm$0.09\% & 6.96$\pm$0.12\% & 7.92$\pm$0.13\%   \\
    \hline
        w/CLN & 7.09$\pm$0.06\% & 7.89$\pm$0.02\% & 7.14$\pm$0.05\% & 8.04$\pm$0.14\%  \\
    \hline
        w/CL & \textbf{7.29$\pm$0.09\%} & \textbf{7.92$\pm$0.04\%} & \textbf{7.28$\pm$0.11\%} & 8.04$\pm$0.12\%  \\
    \hline
    \end{tabular}
    
    \caption{5-shot classification test accuracies on \textit{Fungi} for \textit{Confusable Learning} using the traditional confusion matrix and the proposed soft confusion matrix.}
    \label{table:ablation_count_fungi}
\end{table}{}

    

\subsubsection{Influence of Confusion Matrix Estimation (CME).}

To demonstrate the performance and the efficiency of CME, we implement \textit{Confusable Learning} with the traditional way of confusion matrix calculation, which is performing the meta-learning algorithm on a $K$-way task and calculating confusion matrix with Eq.(~\ref{equ:prob_confusion_matrix}). In fact, the traditional confusion matrix calculation can be seen as a special case of CME, in which $\rho$ is set to $0$ and $N_T^{\text{e}}$ is set to $K$. We denote this setting by attaching a ``w/CLT" behind the name of meta-learning algorithm. 

\begin{table}[!]
\centering
\begin{tabular}{|l|r|r|r|r|r|}  
\hline
 & PN w/CL-1 & PN w/CL-2 & PN w/CL-4  & PN w/CL-8 & PN w/CLT\\
\hline
Accuracy & 73.21\% & 73.58\% & 73.46\% & 73.59\% & 73.60\% \\
\hline
Elapsed Time & 0.073s & 0.146s & 0.289s & 0.570s &1.432s  \\
\hline
GPU Memory & 2463MiB & 2479MiB & 2511MiB & 2567MiB & 12457MiB\\
\hline
\end{tabular}
\caption{5-shot classification test accuracies, averaged elapsed time in each iteration and memory in need for confusion matrix calculation in \textit{Omniglot} (2636-way).}
\label{tab:ablation_cme}
\end{table}

In Table~\ref{tab:ablation_cme}, we compare the results of \textit{Confusable Learning} with the traditional confusion matrix calculation and the result of \textit{Confusable Learning} with our proposed CME. Here, we denote \textit{Confusable Learning} with $M$ steps of CME by attaching a ``w/CL-$M$". The experiment is conducted on a machine with a Tesla P100 GPU. It can be concluded that the CME settings achieve almost the same accuracy with the setting using the traditional confusion matrix calculation but largely decrease the elapsed time and memory requirement. It is noteworthy that \textit{Omniglot} is a small dataset, so we are able to implement  \textit{Confusable Learning} with the traditional confusion matrix calculation easily. When training with a larger dataset like $\textit{fungi}$ and $\textit{Imagenet64x64}$, traditional confusion matrix calculation will require much longer time and larger memory.

\subsection{Parameter Sensitivity Analysis}

\textit{Confusable Learning} contains 5 parameters: $N_D$, $N_T^{\text{e}}$, $N_T^{\text{c}}$, $M$ and $\rho$. Table \ref{tab:ablation_cme} already shows that larger $M$ yields a better result. In this section, we discuss how sensitive the other 4 parameters are.

\begin{figure}[!]
\centering
\subfigure[$\rho$ vs. accuracy]{
\centering
\includegraphics[width=2.7cm]{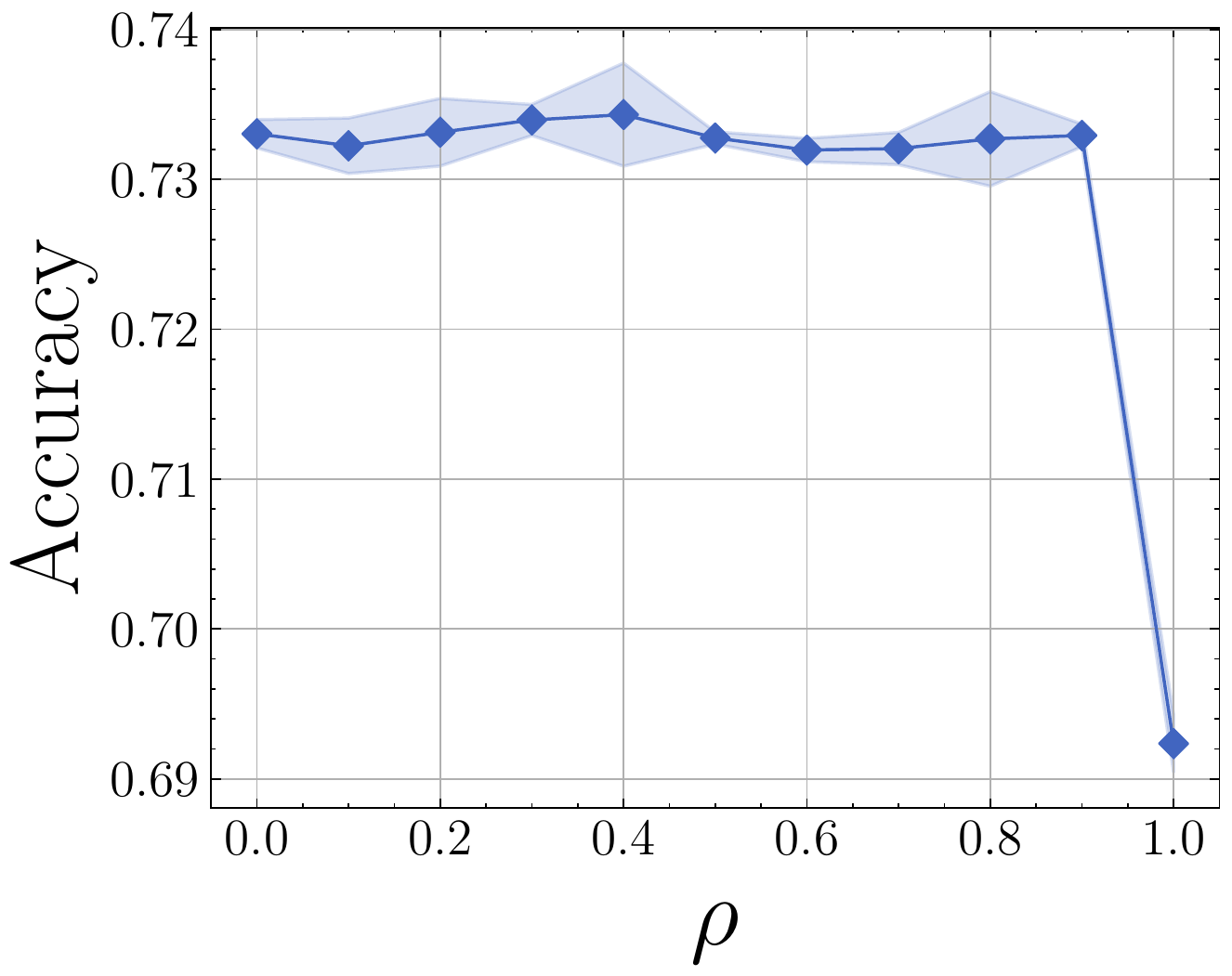}
\label{fig:sensitivity:rho}
}
\subfigure[$N_D$ vs. accuracy]{
\centering
\includegraphics[width=2.7cm]{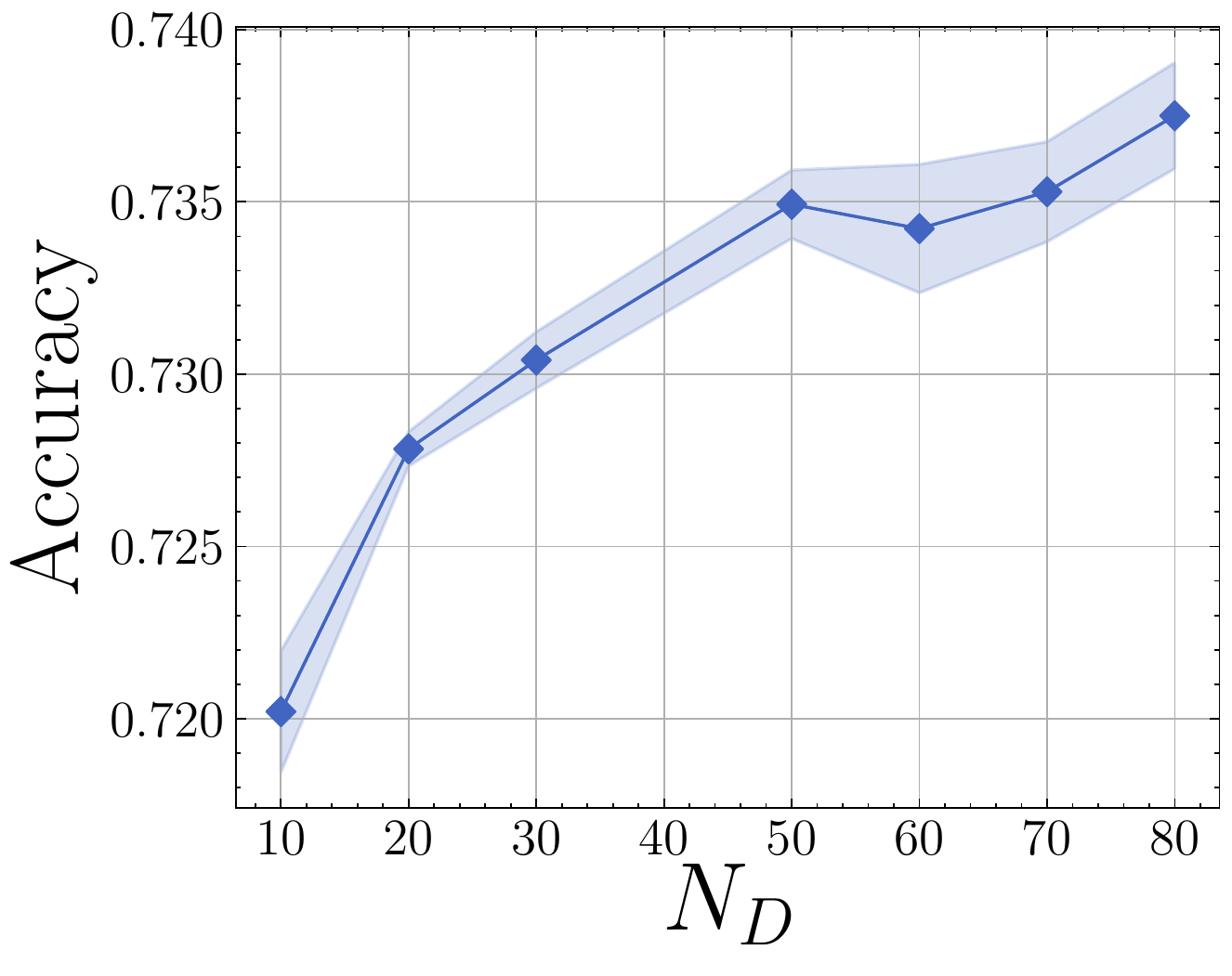}
\label{fig:sensitivity:nd}
}
\subfigure[$N_T^{\text{e}}$ vs. accuracy]{
\centering
\includegraphics[width=2.7cm]{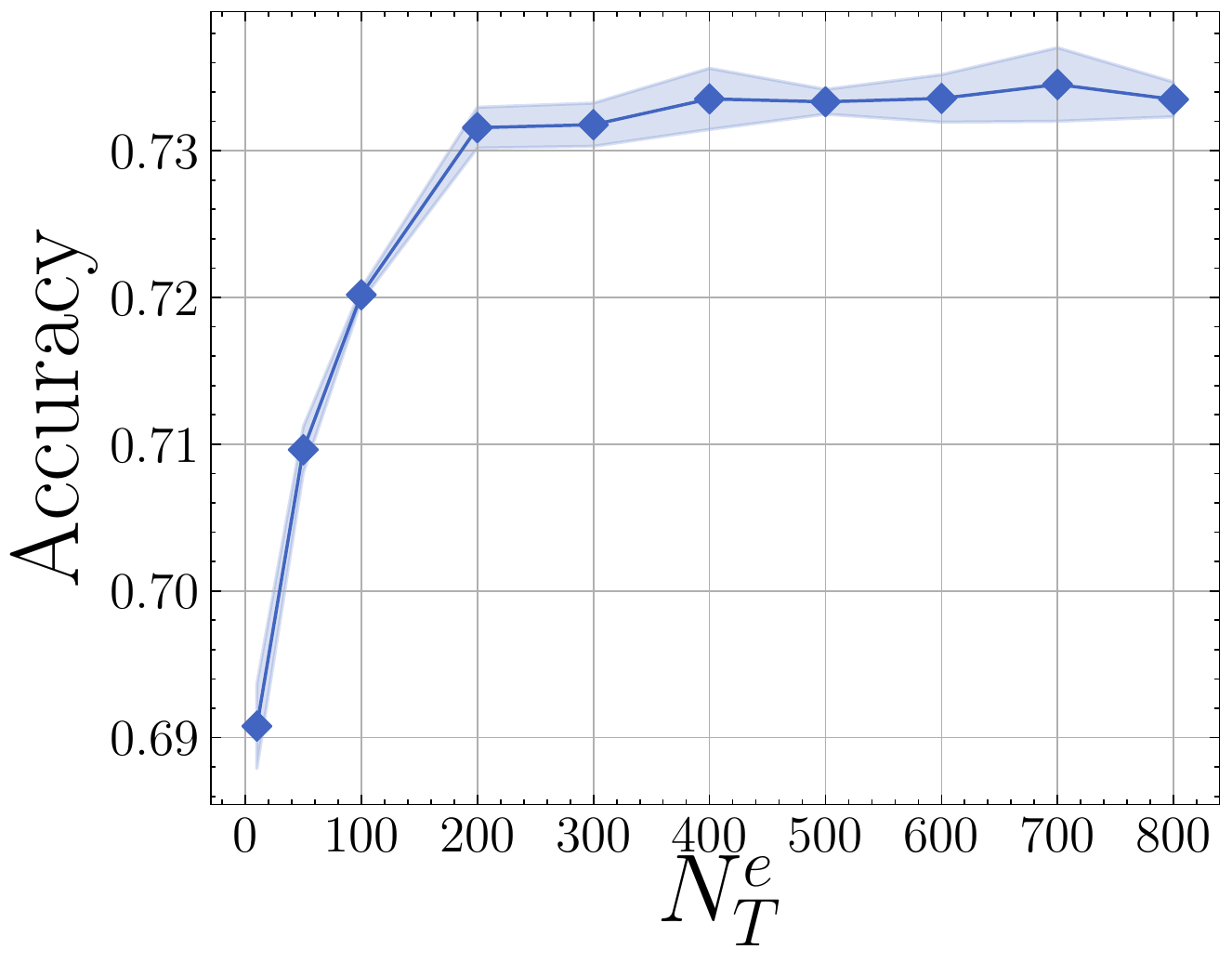}
\label{fig:sensitivity:net}
}
\subfigure[$N_T^{\text{c}}$ vs. accuracy]{
\centering
\includegraphics[width=2.7cm]{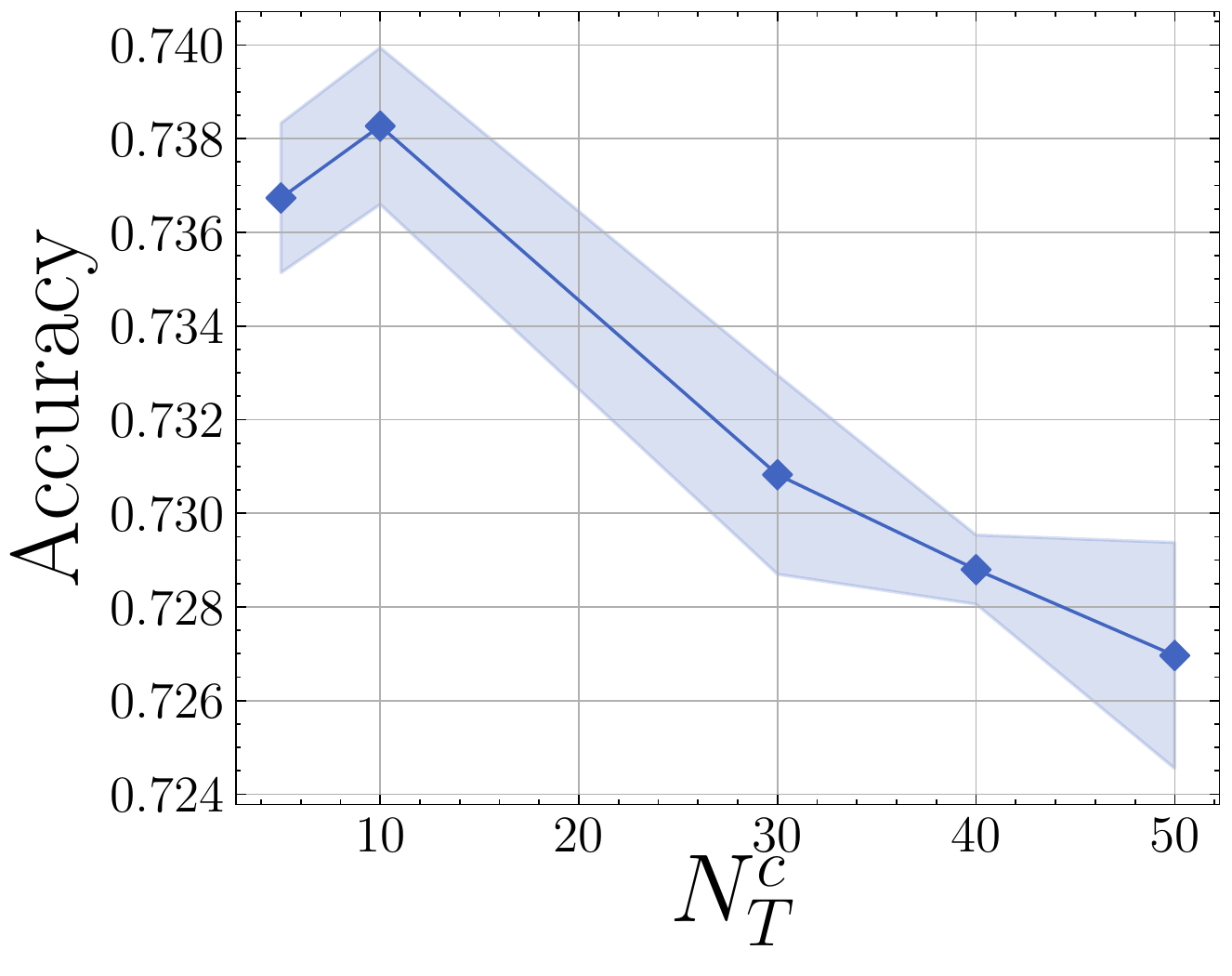}
\label{fig:sensitivity:nct}
}
\caption{Accuracy of PN w/CL vs. parameters in \textit{Omniglot}.}
\label{fig:sensitivity}
\end{figure}

Based on the parameters we use in \textit{Omniglot} PN w/CL experiment in Section \ref{experiment setup}, we adjust each of the 4 parameters at a time. The results are shown in Figure~\ref{fig:sensitivity}. Performance of the model is very stable with $\rho$ changing from $0$ to $0.9$. It is not surprising that the accuracy drops below 70\% when $\rho$ is set to $1$. In such a case, the confusion matrix will not be updated and thus, \textit{Confusable Learning} can not obtain any useful information from the confusion matrix. Increasing $N_D$ and $N_T^{\text{e}}$ always help improve accuracy but requires longer elapsed time and more memory. When increasing $N_T^{\text{c}}$, as shown in \ref{fig:sensitivity:nct}, accuracy firstly increases and then decreases. The optimal $N_T^{\text{c}}$ is about $10$. Regardless of this observation, the accuracy is over 73\% within a large range of $N_T^{\text{c}}$, significantly higher than the accuracy of the PN model without $Confusable Learning$ reported in Table~\ref{tab:acc}. It can be concluded that \textit{Confusable Learning} can yield a great performance without any elaborate tuning. 


\section{Conclusion}

We have presented an approach to locate and learn from confusable classes in large-class few-shot classification problem. We show significant gains on top of multiple meta-learning methods, achieving state-of-the-art performance on three challenging datasets. Future work will involve constructing better confusion tasks to learn confusable classes better.


\begin{thebibliography}{8}

\bibitem{finn2017model}
Finn, C., Abbeel, P., Levine, S.: Model-agnostic meta-learning for fast adaptation of deep networks. In: International Conference on Machine Learning (ICML), pp. 1126-1135. (2017)

\bibitem{snell2017prototypical}
Snell, J., Swersky, K., Zemel, R.: Prototypical networks for few-shot learning. In: Advances in Neural Information Processing Systems (NeurIPS), pp. 4077-4087. (2017)

\bibitem{wang2019few}
Wang, Y., Yao, Q.: Few-shot learning: A survey. arXiv preprint \url{arXiv:1904.05046}. (2019)

\bibitem{vinyals2016matching}
Vinyals, O., Blundell, C., Lillicrap, T., Wierstra, D.: Matching networks for one shot learning. In: Advances in Neural Information Processing Systems (NeurIPS), pp. 3630-3638. (2016)

\bibitem{chen2018a}
Chen, W., Liu, Y., Zsolt, K., Wang, Y., Huang, J.: A Closer Look at Few-shot Classification. In: The International Conference on Learning Representations (ICLR),(2019)

\bibitem{NIPS2018_7504}
Zhang, R., Che, T., Ghahramani, Z., Bengio, Y., Song, Y.: MetaGAN: An Adversarial Approach to Few-Shot Learning. In: Advances in Neural Information Processing Systems (NeurIPS), pp. 2365-2374. (2018)

\bibitem{NIPS2018_7352}
Oreshkin, B., Rodr\'{\i}guez L\'{o}pez, P., Lacoste, A.: TADAM: Task dependent adaptive metric for improved few-shot learning. In:Advances in Neural Information Processing Systems (NeurIPS), pp. 721-731. (2018)

\bibitem{liu2018learning}
Liu, Y., Lee, J., Park, M., Kim, S., Yang, E., Hwang, S., Yang, Y.: Learning to propagate labels: transductive propagation network for few-shot learning. In: Conference on Computer Vision and Pattern Recognition (CVPR), (2018)

\bibitem{pmlr-v80-franceschi18a}
Franceschi, L., Frasconi, P., Salzo, S., Grazzi, R., Pontil, M.: Bilevel Programming for Hyperparameter Optimization and Meta-Learning. In: International Conference on Machine Learning (ICML), (2018)

\bibitem{hariharan2017low}
Hariharan, B., Girshick, Ross.: Low-shot visual recognition by shrinking and hallucinating features. In: International Conference on Computer Vision (ICCV), pp. 3018-3027. (2017)

\bibitem{Li_2019_CVPR}
Li, A., Luo, T., Lu, Z., Xiang, T., Wang, L.: Large-Scale Few-Shot Learning: Knowledge Transfer With Class Hierarchy. In: Conference on Computer Vision and Pattern Recognition (CVPR), pp. 7212-7220. (2019)

\bibitem{deng2010does}
Deng, J., Berg, A., Li, K., Li, F.: What does classifying more than 10,000 image categories tell us? In European Conference on Computer Vision (ECCV), pp. 71-84. (2010)


\bibitem{wang2018low}
Wang, Y., Girshick, R., Hebert, M., Hariharan, B.: Low-shot learning from imaginary data. In: Conference on Computer Vision and Pattern Recognition (CVPR), pp. 7278-7286. (2018)

\bibitem{bertinetto2018metalearning}
Bertinetto, L., Henriques, J., Torr, P., Vedaldi, A.: Meta-learning with differentiable closed-form solvers. In: The International Conference on Learning Representations (ICLR), (2019)


\bibitem{lake2015human}
Lake, B., Salakhutdinov, R., Tenenbaum, J.: Human-level concept learning through probabilistic program induction. Science 350(6266), 1332--1338 (2015)


\bibitem{imagenet_cvpr09}
Deng, J., Dong, W., Socher, R., Li, L., Li, Kai., Li, F.: ImageNet: A Large-Scale Hierarchical Image Database. In: Conference on Computer Vision and Pattern Recognition (CVPR), pp. 248-255. (2009)


\bibitem{Ren2018Meta}
Ren, M., Ravi, S., Triantafillou, E., Snell, J., Swersky, K., Tenenbaum, J., Larochelle, H., Zemel, R.: Meta-Learning for Semi-Supervised Few-Shot Classification. In: The International Conference on Learning Representations (ICLR), (2018)

\bibitem{wichrowska2017learned}
Wichrowska, O., Maheswaranathan, N., Hoffman, M., Colmenarejo, S., Denil, M., de Freitas, N., Sohl-Dickstein, J.: Learned optimizers that scale and generalize. In: International Conference on Machine Learning (ICML), pp. 3751-3760. (2017)

\bibitem{koch2015siamese}
Koch, G., Zemel, R., Salakhutdinov, R.: Siamese neural networks for one-shot image recognition. In: International Conference on Machine Learning Deep Learning Workshop (ICML), (2015)



\bibitem{garcia2018fewshot}
Garcia, V., Bruna, E.: Few-Shot Learning with Graph Neural Networks. In: The International Conference on Learning Representations (ICLR), (2018)

\bibitem{liu2019prototype}
Liu, L., Zhou, T., Long, G., Jiang. J., Yao, L., Zhang, C.: Prototype propagation networks (PPN) for weakly-supervised few-shot learning on category graph. In International Joint Conferences on Artificial Intelligence (IJCAI), (2019)

\bibitem{liu2019learning}
Liu, L., Zhou, T., Long, G., Jiang. J., Zhang, C.: Learning to propagate for graph meta-learning. In Advances in Neural Information Processing Systems (NeurIPS), pp.1037-1048. (2019)


\bibitem{Gupta2014Training}
Gupta, R., Bengio, S., Weston, J.: Training Highly Multiclass Classifiers. Journal of Machine Learning Research (JMLR) 15(1), 1461-1492 (2014)



\bibitem{he2016deep}
He, K., Zhang, X., Ren, S., Sun, J.: Deep residual learning for image recognition. In: Conference on Computer Vision and Pattern Recognition (CVPR), pp. 770-778. (2016)










\bibitem{sun2019meta}
Sun, Q., Liu, Y., Chua, T., Schiele, B.: Meta-transfer learning for few-shot learning. In Conference on Computer Vision and Pattern Recognition (CVPR), pp. 403-412. (2019)

\bibitem{inproceedings}
Griffin, G., Perona, P.: Learning and using taxonomies for fast visual categorization. In: Conference on Computer Vision and Pattern Recognition (CVPR), pp. 1-8. (2008)



\bibitem{chrabaszcz2017downsampled}
Chrabaszcz, P., Loshchilov, I., Hutter, F., A Downsampled Variant of ImageNet as an Alternative to the CIFAR datasets. arXiv preprint \url{arXiv:1707.08819}. (2017)

\bibitem{normalizedconfusionmatrix}
Giannakopoulos, T., Pikrakis, A.: Introduction to audio analysis: a MATLAB® approach. Academic Press. Academic Press. (2014)




\end{thebibliography}
\end{document}